%% file: main.tex
\documentclass[10pt,twocolumn,letterpaper]{article}

\usepackage[pagenumbers]{cvpr} %

\input{preamble}

\title{\paperTitleNewLine{}}

\author{
Ho Kei Cheng\textsuperscript{1}\thanks{Work done during an internship at Sony AI.}\hspace{2em}
Masato Ishii\textsuperscript{2}\hspace{2em}
Akio Hayakawa\textsuperscript{2}\hspace{2em}
Takashi Shibuya\textsuperscript{2} 
\\
Alexander Schwing\textsuperscript{1}\hspace{2em}
Yuki Mitsufuji\textsuperscript{2,3}
\\
\textsuperscript{1}University of Illinois Urbana-Champaign\hspace{2em} \textsuperscript{2}Sony AI\hspace{2em}
\textsuperscript{3}Sony Group Corporation 
\\
{\tt\scriptsize
\{hokeikc2,aschwing\}@illinois.edu, \{masato.a.ishii,akio.hayakawa,takashi.tak.shibuya,yuhki.mitsufuji\}@sony.com}
}

\begin{document}
\maketitle
\input{sec/00-abstract}
\input{sec/01-intro}

\input{sec/02-related}
\input{sec/03-method}
\input{sec/04-expr}

\input{sec/05-conclusion}

{
    \small
    \bibliographystyle{ieeenat_fullname}
    \bibliography{main}
}

\onecolumn
\clearpage
\input{sec/10-appendix}

\end{document}

%% file: preamble.tex
\def\paperID{2225} 
\def\confName{CVPR}
\def\confYear{2025}
\definecolor{cvprblue}{rgb}{0.21,0.49,0.74}

\def\paperTitleNewLine{MMAudio: Taming Multimodal Joint Training \\for High-Quality Video-to-Audio Synthesis}
\def\paperTitle{MMAudio: Taming Multimodal Joint Training for High-Quality Video-to-Audio Synthesis}

\usepackage[accsupp]{axessibility}

\usepackage[utf8]{inputenc} %
\usepackage[T1]{fontenc}    %
\usepackage[pagebackref,breaklinks,colorlinks,allcolors=cvprblue]{hyperref}
\usepackage{url}            %
\usepackage{booktabs}       %
\usepackage{amsfonts}       %
\usepackage{nicefrac}       %
\usepackage{microtype}      %
\usepackage[dvipsnames]{xcolor}         %
\usepackage{notoccite}
\usepackage{titletoc}

\usepackage{colortbl}
\usepackage{graphicx,overpic}
\usepackage[export]{adjustbox}
\usepackage{amsmath, amssymb}
\usepackage{multirow}
\usepackage[super]{nth}
\usepackage{comment}
\usepackage{nicematrix}

\usepackage{enumitem}
\usepackage{caption}

\usepackage{titlesec}

\titleformat{\section}
{\normalfont\large\bfseries}{\thesection.~}{0pt}{}{}
\titleformat{\subsection}
{\normalfont\large\bfseries}{\thesubsection.~}{0pt}{}{}
\titleformat{\subsubsection}
{\normalfont\normalsize\bfseries}{\thesubsubsection.~}{0pt}{}{}
\titleformat{\paragraph}[runin]
{\normalfont\normalsize\bfseries}{\theparagraph\vspace{1em}}{}{}

\titlespacing*{\section}{0pt}{2.0ex plus .5ex minus .5ex}{7pt plus 2px minus 2px}
\titlespacing*{\subsection}{0pt}{1.0ex plus .5ex minus .5ex}{5pt plus 2px minus 2px}
\titlespacing*{\subsubsection}{0pt}{.5ex plus .5ex minus .5ex}{3pt plus 1pt minus 1pt}
\titlespacing*{\paragraph}{0em}{.5ex plus .5ex minus .3ex}{1em}

\usepackage{tikz}
\usepackage{pgfplots}
\usepackage{pgfplotstable}
\pgfplotsset{compat=1.13}
\usetikzlibrary{pgfplots.groupplots}
\usetikzlibrary{calc,arrows}

\newcommand\pmnum[1]{\small$\pm$#1}

\newcommand*\conditions{\mathbf{C}}

\newcommand*\featsync{F_{\mathit{syn}}}
\newcommand*\featvis{F_v}
\newcommand*\feattext{F_t}
\newcommand*\feataudio{x}

\newcommand*\fdvgg{FD\(_{\text{VGG}}\)}
\newcommand*\fdpann{FD\(_{\text{PANNs}}\)}
\newcommand*\fdpasst{FD\(_{\text{PaSST}}\)}
\newcommand*\ispann{IS}

\newcommand*\klpann{KL\(_{\text{PANNs}}\)}
\newcommand*\klpasst{KL\(_{\text{PaSST}}\)}

\usepackage{tabularx,stackengine,collcell}
\let\endminwd\relax
\newcolumntype{L}[1]{>{\collectcell\xminwd l{#1}}l<{\endminwd\endcollectcell}}
\newcolumntype{C}[1]{>{\collectcell\xminwd c{#1}}c<{\endminwd\endcollectcell}}
\newcolumntype{R}[1]{>{\collectcell\xminwd r{#1}}r<{\endminwd\endcollectcell}}
\def\minwd#1#2#3\endminwd{\stackengine{0pt}{#3}{\rule{#2}{0pt}}{O}{#1}{F}{F}{L}}
\newcommand\xminwd[1]{\minwd#1}

\hypersetup{
	colorlinks,
	linkcolor={red!80!black},
	citecolor={cvprblue},
	urlcolor={cvprblue},
    pdftitle={\paperTitle{}},
}

\definecolor{defaultColor}{RGB}{230, 244, 252}

\newcommand{\beginsupplement}{
    \appendix
	\setcounter{table}{0}
	\renewcommand{\thetable}{A\arabic{table}}%
	\setcounter{figure}{0}
	\renewcommand{\thefigure}{A\arabic{figure}}%
	\setcounter{equation}{0}
	\renewcommand{\theequation}{A\arabic{equation}}
}

%% file: sec/00-abstract.tex
\begin{abstract}
We propose to synthesize high-quality and synchronized audio, given video and optional text conditions, using a novel multimodal joint training framework (\textbf{MMAudio}).
In contrast to single-modality training conditioned on (limited) video data only, MMAudio is jointly trained with larger-scale, readily available text-audio data to learn to generate semantically aligned high-quality audio samples.
Additionally, we improve audio-visual synchrony with a conditional synchronization module that aligns video conditions with audio latents at the frame level.
Trained with a flow matching objective, MMAudio achieves new video-to-audio state-of-the-art among public models in terms of audio quality, semantic alignment, and audio-visual synchronization, while having a low inference time (1.23s to generate an 8s clip) and just 157M parameters.
MMAudio also achieves surprisingly competitive performance in text-to-audio generation, showing that joint training does not hinder single-modality performance.
Code, models, and demo are available at: {\href{https://hkchengrex.github.io/MMAudio}{\nolinkurl{hkchengrex.github.io/MMAudio}.}}

\end{abstract}

%% file: sec/01-intro.tex
\section{Introduction}
\label{sec:intro}

We are interested in \emph{Foley}, \ie, for a given video we want to synthesize ambient sound (\eg, rain, river flow) and sound effects induced by visible events (\eg, a dog barks, a racket hits a tennis ball). 
Note, Foley does not focus on synthesizing background music or human speech, which is often added in post-processing.
Importantly, Foley requires to 
synthesize convincing high-quality audio that is 1) semantically and 2) temporally aligned to an input video.
For semantic alignment, methods need to understand scene contexts and their association with audio -- the visual concept of rain should be associated with the sound of splashing raindrops. 
For temporal alignment, 
methods need to understand audio-visual synchrony as humans can perceive audio-visual misalignment as slight as 25~ms~\cite{petrini2009multisensory}.
Inspired by the efficacy of training data scaling demonstrated by recent works~\cite{kaplan2020scaling,radford2021learning,esser2024scaling}, we pursue a data-driven approach to %
synthesize high-quality audio that respects these two types of alignment.

\begin{figure}
    \centering
    \includegraphics[width=\linewidth]{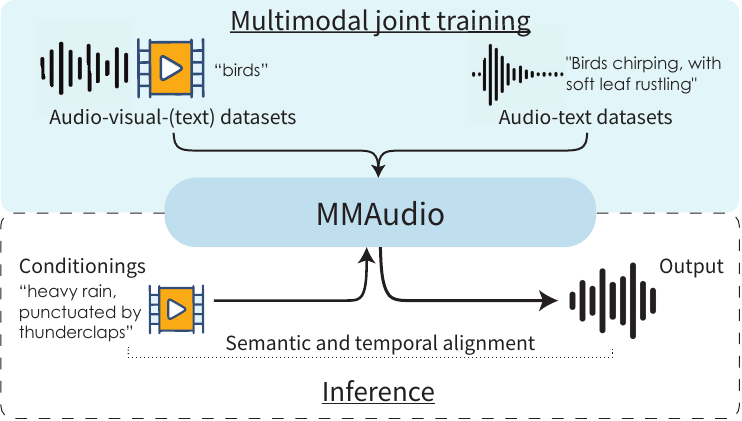}
    \caption{In addition to training on audio-visual-(text) datasets, we perform multimodal joint training with high-quality, abundant audio-text data which enables effective data scaling. 
    At inference, MMAudio generates conditions-aligned audio with video and/or text guidance.}
    \label{fig:teaser}
\end{figure}

Current state-of-the-art video-to-audio methods either train only on audio-visual data~\cite{chen2020vggsound} from scratch~\cite{wang2024frieren} or train new ``control modules''~\cite{zhang2024foleycrafter,jeong2024read,wang2024v2a,xing2024seeing} for pretrained text-to-audio models on audio-visual data.
The former is limited by the amount of available training data: the most commonly used audio-visual dataset VGGSound~\cite{chen2020vggsound} contains only around 550 hours of videos.
Audio-visual data are expensive to curate at a large scale, as in-the-wild videos from the Internet 
1) contain music and speech\footnote{In AudioSet~\cite{gemmeke2017audio} (collected from YouTube), 82.48\% of videos contain either human speech or music according to the provided audio labels.}, which has limited utility for training a general Foley model, and
2) contain non-diegetic~\cite{stilwell200711} sounds such as background music or sound effects added in post-processing, again unsuitable for a Foley model.
The latter, \ie, finetuning pretrained text-to-audio models (with added control modules) on audio-visual data, enables models to benefit from audio-generation knowledge obtained from larger-scale audio-text data (\eg, 
WavCaps~\cite{mei2024wavcaps} at 7,600 hours).
However, adding control modules to pretrained text-to-audio models complicates the network architecture and limits the design space.
It is also unclear whether pretrained text-to-audio models have sufficient degrees of freedom to support all video-to-audio scenarios, compared to training from scratch.

To avoid these limitations, we propose a \emph{multimodal joint training} paradigm (\Cref{fig:teaser}) that jointly considers video, audio, \emph{and} text in a single multimodal transformer network and masks missing modalities during training.
This allows us to train from scratch on both audio-visual and audio-text datasets with a simple end-to-end framework.
Jointly training on large multimodal datasets enables a unified semantic space and exposes the model to more data for learning the distribution of natural audio.
Empirically, with joint training, we observe a significant relative improvement in audio quality (10\% lower Fr\'{e}chet Distance~\cite{kilgour2018fr} and 15\% higher Inception Score~\cite{salimans2016improved}), semantic alignment (4\% higher ImageBind~\cite{girdhar2023imagebind} score), and temporal alignment (14\% better synchronization score).

To further improve temporal alignment, 
we introduce a conditional synchronization module that uses high frame-rate visual features (extracted from a self-supervised audio-visual desynchronization detector~\cite{iashin2024synchformer}) and 
operates in the space of scales and biases of adaptive layer normalization (adaLN) layers~\cite{perez2018film}, leading to accurate synchronization (50\% relative improvement in synchronization score).

In summary, we first propose MMAudio, a \textbf{multimodal joint training} paradigm for video-to-audio. It enables accessible data scaling and cross-modal understanding, significantly improving audio quality and semantic alignment. 
We also propose a \textbf{conditional synchronization module} which enables more precise audio-visual synchrony.
We train MMAudio on openly accessible datasets and scale it to two audio sampling rates (16kHz and 44.1kHz) and three model sizes (157M, 621M, 1.03B), with the smallest model already achieving new state-of-the-art performance in video-to-audio synthesis among public models.
Surprisingly, our multimodal approach also achieves competitive performance in text-to-audio generation compared to dedicated text-to-audio methods, showing that joint training does not hinder single-modality performance.

%% file: sec/02-related.tex
\section{Related Works}
\label{sec:related}

\paragraph{Semantic alignment.}
Semantic alignment between audio and video is learned by training on paired audio-visual data with a generation objective~\cite{iashin2021taming, du2023conditional, mei2023foleygen,wang2024v2a,pascual2024masked,su2023physics,chen2020generating,yang2024draw} or a contrastive objective~\cite{luo2024diff}.
To further understand audio semantics, we additionally train on paired audio-text data.
We argue that the semantic understanding learned from audio-text pairs can be transferred to video-text pairs, as joint training leads to a shared semantic space (similar to ImageBind~\cite{girdhar2023imagebind} and LanguageBind~\cite{zhu2023languagebind}) and enables the network to learn richer semantics from more diverse data.

\paragraph{Temporal alignment.}
Besides learning temporal alignment directly from audio-visual pairs, some recent works first learn from videos a separate model to predict handcrafted proxy features such as audio onsets~\cite{zhang2024foleycrafter,ren2024sta}, energy~\cite{jeong2024read,huang2024rhythmic}, or root-mean-square of waveforms~\cite{lee2024video,chung2024t}.
We deviate from these handcrafted features 
and directly learn alignment from the deep feature embeddings of a pre-trained self-supervised desynchronization detector Synchformer~\cite{iashin2024synchformer}, allowing a more nuanced interpretation of the input signal.
A recent work V-AURA~\cite{viertola2024temporally} also uses Synchformer~\cite{iashin2024synchformer} for synchronization in an autoregressive framework.
However, \cite{viertola2024temporally} does not perform multimodal training on text and has a short context window (2.56s) while we produce longer-term (8-10s) temporally consistent generations.
In terms of improving positional embeddings for temporal alignment, \citet{mei2024towards} concurrently propose to subsample high-frequency (audio) positional embeddings, while we propose to scale up the frequencies of low-frequency (visual) positional embeddings -- the effects are identical when the higher frequency is an integer multiple of the lower one.

\paragraph{Multimodal conditioning.}
The most common way to support multimodal conditioned generation is to add ``control modules'' that inject 
visual features
to a pretrained text-to-audio network~\cite{zhang2024foleycrafter,jeong2024read,ren2024sta,lee2024video,mo2024text,huang2024rhythmic,haji2024av}.
However, this increases the number of parameters. %
Besides, as the text modality is fixed during video-to-audio training, it becomes more challenging to learn a joint semantic space -- the video modality needs to bind to the semantics of text instead of both modalities learning to cooperate. 
In contrast, we train all modalities simultaneously in our multimodal training formulation to learn joint semantics and enable omnidirectional feature sharing among modalities.
Alternatively, to align different modalities without training, Seeing-and-Hearing~\cite{xing2024seeing} uses a pretrained text-to-audio model and performs gradient ascent on an alignment score (\ie, ImageBind~\cite{girdhar2023imagebind}) at test-time.
We note that the test time optimization is slow and sometimes results in low-quality and temporally misaligned output.
Indeed, our model is faster at test time and more consistently produces synchronized audio.
Concurrent to our work, VATT~\cite{liu2024tellhearvideo} and MultiFoley~\cite{chen2024video} explore jointly trained multimodal conditioning. 
VATT~\cite{liu2024tellhearvideo} uses both video and text to generate audio but always requires video conditions during training.
MultiFoley~\cite{chen2024video} is formulated similarly to ours, but MMAudio uses much higher frame-rate visual features (24 FPS, while MultiFoley uses 8 FPS features), which leads to significantly better audio-visual synchronization.

\paragraph{Multimodal generation.}
Related to multimodal conditioning, multimodal generation models produce samples composed of multiple modalities (\eg, video \emph{and} audio).
Multimodal generation is more challenging and existing approaches~\cite{kim2024versatile,ruan2022mm,tang2024any,tang2024codi} are not yet competitive with dedicated video-to-audio models.
In this work, we focus on multimodal conditioned audio generation.
We believe our multimodal formulation and architecture serve as a foundation for future work in multimodal generation.

%% file: sec/03-method.tex
\begin{figure*}[t]
    \centering
    \includegraphics[width=0.9\linewidth]{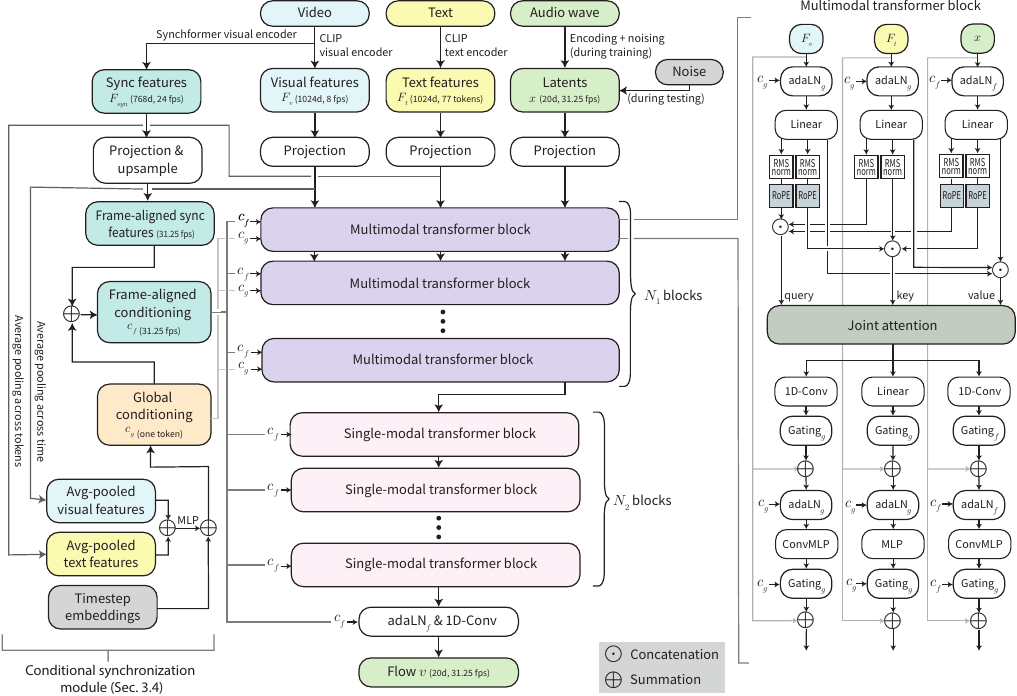}
    \caption{
    Overview of the MMAudio flow-prediction network.
    Video conditions, text conditions, and audio latents jointly interact in the multimodal transformer network.
    A synchronization model (\Cref{sec:cond_sync}) injects frame-aligned synchronization features for precise audio-visual synchrony.
    }
    \label{fig:overview}
\end{figure*}

\section{MMAudio}
\label{sec:method}

\subsection{Preliminaries}
\label{sec:preliminaries}

\paragraph{Conditional flow matching.}
We use the conditional flow matching objective~\cite{lipman2022flow,tong2023improving} for generative modeling and refer readers to~\cite{tong2023improving} for details.
In short, at test time, to generate a sample, we randomly draw noise \(x_0\) from the standard normal distribution and use an ODE solver to numerically integrate from time \(t=0\) to time \(t=1\) following a learned time-dependent conditional velocity vector field 
\(v_\theta(t, \conditions{}, x): [0, 1]\times \mathbb{R}^C \times \mathbb{R}^d \to \mathbb{R}^d\), where \(t\) is the timestep, \(\conditions{}\) is the condition (\eg, video and text), and \(x\) is a point in the vector field. We represent the velocity vector field via a deep net parameterized by \(\theta\). %

At training time, we find $\theta$ by considering the conditional flow matching objective
\begin{equation}
    \mathbb{E}_{t, q(x_0), q(x_1, \conditions{})} \lVert v_\theta(t, \conditions{}, x_t) - u(x_t|x_0, x_1) \rVert^2, 
\label{eq:cfm_objective}
\end{equation}
where \(t\in[0, 1]\), \(q(x_0)\) is the standard normal distribution, and \(q(x_1, \conditions{})\) samples from the training data.
Further,  
\begin{equation}
    x_t = tx_1 + (1-t)x_0
\end{equation}
defines a linear interpolation path between noise and data, and
\begin{equation}
    u(x_t|x_0, x_1) = x_1 - x_0
\end{equation}
denotes its corresponding flow velocity at \(x_t\). 

\paragraph{Audio encoding.}
For computational efficiency, we model the generative process in a latent space, following the common practice~\cite{liu2023audioldm,wang2024frieren}.
For this, we first transform audio waveforms via Short-Time Fourier Transform (STFT) and extract mel spectrograms~\cite{stevens1937scale}, which are then encoded by a pretrained variational autoencoder (VAE)~\cite{vae} into latents~$x_1$.
During testing, the generated latents are decoded by the VAE into spectrograms which are then vocoded by a pretrained vocoder~\cite{lee2022bigvgan} into audio waveforms.

\subsection{Overview}

Following conditional flow matching, at test time, we numerically integrate noise $x_0$ from $t=0$ to $t=1$ following a flow $v_\theta$, which was learned at training time by optimizing \cref{eq:cfm_objective}. 
Numerical integration at test time arrives at a latent $x_1$ that is decoded to audio preferably of high-quality and preferably semantically and temporally aligned to the video and text conditions. 

To estimate a flow $v_\theta$ for the current latent $x$, MMAudio operates on the video/text conditions and the flow timestep $t$.
\Cref{fig:overview} illustrates our network architecture. 
To combine inputs from different modalities, MMAudio consists of a series of (\(N_1\)) {\bf multimodal transformer} blocks~\cite{esser2024scaling} with visual/text/audio branches, followed by a series of (\(N_2\)) audio-only transformer blocks~\cite{flux}.
Additionally, for audio-visual synchrony, we devise a {\bf conditional synchronization module} that extracts and integrates into the generation process high frame rate (24 frames per second (fps)) visual features for temporal alignment.
Next, we describe both components in detail.

\subsection{Multimodal Transformer}
Core to our approach is the desire to model the interactions between video, audio, and text modalities.
For this purpose, we largely adopt the MM-DiT block design from SD3~\cite{esser2024scaling} and introduce two new components for temporal alignment: aligned RoPE positional embeddings for aligning sequences of different frame rates and 1D convolutional MLPs (ConvMLPs) for capturing local temporal structure.
\Cref{fig:overview} (right) illustrates our block design.
Note, we also include a sequence of audio-only single-modality blocks following FLUX~\cite{flux}, implemented by simply removing the streams of the two other modalities (\ie, the joint attention becomes a self-attention).
Compared to considering all modalities at every layer, this design allows us to build a deeper network with the same parameter count and compute without sacrificing multimodality.
This multimodal architecture allows the model to selectively attend to and focus on different modalities depending on the inputs, thus enabling effective joint training on both audio-visual and audio-text data.
Next, we describe the feature representation in our network and then the core components of our block design.

\paragraph{Representations.}
We represent all the features as one-dimensional tokens.
Note, we deliberately do not use any absolute position encoding which allows us to generalize to different durations at test time.
Thus, we specify the temporal sequences in terms of frame rates to determine the number of tokens for a given duration.
The visual features \(\featvis\) (one token per frame, at 8 fps) and text features \(\feattext\) (77 tokens) are extracted from CLIP~\cite{radford2021learning} as 1024d features.
The audio latents \(\feataudio\) are in the VAE latent space (\Cref{sec:preliminaries}), at 31.25 fps as 20d latents by default.
The synchronization features \(\featsync\) are extracted with Synchformer~\cite{iashin2024synchformer} at 24 fps as 768d features, which we will detail in~\Cref{sec:cond_sync}.
Note, except for the text tokens, all other tokens follow the same temporal ordering, albeit at different frame rates.
After the initial ConvMLP/MLP layers, all features will be projected to the hidden dimension $h$.

\paragraph{Joint attention.}
These tokens from different modalities communicate via joint attention (\Cref{fig:overview}, right).
Following~\cite{esser2024scaling}, we concatenate the query, key, and value representations from the three different modalities and apply 
scaled dot product attention~\cite{vaswani2017attention}.
The output is split into three modalities, following the input partition.
We refer readers to~\cite{esser2024scaling} for details.
We note that joint attention alone does not capture temporal alignment which we will address next.

\paragraph{Aligned RoPE position embedding.}
For audio-visual synchrony, precise temporal alignment is crucial.
As typical in transformers~\cite{vaswani2017attention}, we adopt positional embeddings to inform the attention layers of time.
Specifically, we apply RoPE~\cite{su2024roformer} embeddings on the queries and keys in the visual and audio streams before joint attention (\Cref{fig:overview}).
Note that we do not apply it to the text stream since it does not follow the temporal order of video or audio.
Further, since the frame rates do not align (8 fps for the visual stream, 31.25 fps for the audio stream), we scale the frequencies of the positional embeddings in the visual stream proportionally%
, \ie, by \(31.25/8\).
We visualize the default (non-aligned) RoPE and our proposed aligned RoPE in~\Cref{app:fig:aligned_rope}.
We note that these aligned embeddings are beneficial yet insufficient for good synchrony. Therefore, we introduce an additional synchronization module, which we discuss in \Cref{sec:cond_sync}.

\paragraph{ConvMLP.}
To better capture local temporal structure, we use ConvMLPs rather than MLPs in the visual and audio streams.
Concretely, our ConvMLP uses 1D convolutions (kernel size \(=3\) and padding \(=1\)) rather than linear layers.
Again, this change is not made to the text stream since it does not follow the temporal order of video or audio.

\paragraph{Global conditioning.}
Global conditioning injects global features into the network through scales and biases in adaptive layer normalization layers (adaLN)~\cite{perez2018film}.
First, we compute a global conditioning vector \(c_g\in\mathbb{R}^{1\times h}\) 
shared across all transformer blocks from the Fourier encoding~\cite{vaswani2017attention} of the flow timestep, the average-pooled visual features, and the average-pooled text features (\Cref{fig:overview}).
Then, each adaLN layer modulates its input \(y\in\mathbb{R}^{L\times h}\) (\(L\) is the sequence length) with the global condition \(c_g\) as follows: 
\begin{equation}
    \text{adaLN}_g(y, c_g) = \text{LayerNorm}(y) \cdot \mathbf{1}\mathbf{W}_\gamma(c_g) + \mathbf{1}\mathbf{W}_\beta(c_g).
\label{eq:global_condition}
\end{equation}
Here, \(\mathbf{W}_\gamma, \mathbf{W}_\beta\) are MLPs, and $\mathbf{1}$ is a $L\times1$ all-ones matrix, which ``broadcasts'' the scales and biases to match the sequence length $L$ -- such that the same condition is applied to all tokens in the sequence (hence global).
Next, we describe how we design position-dependent conditions for precise audio-visual synchronization.

\subsection{Conditional Synchronization Module}
\label{sec:cond_sync}
We develop a token-level conditioning to further improve audio-visual synchrony.
While the visual and audio streams already communicate via cross-modality attention layers, these layers aggregate features via a soft distribution, which we found to hamper precision.
To address this issue, we first extract high frame rate (24 fps) features (\(\featsync\)) from the input video using the visual encoder of Synchformer~\cite{iashin2024synchformer}.
We use Synchformer because it is trained in a self-supervised manner to detect temporal misalignment between video and audio data, which we hypothesize will yield visual features relevant to audio events and hence benefit synchronization.%
We find the frame-aligned conditioning \(c_f\in\mathbb{R}^{L\times h}\) via
\begin{equation}
    c_f = \text{Upsample}\left( \text{ConvMLP}\left( \featsync \right)  \right) + \mathbf{1}c_g.
\end{equation}
Upsample 
uses nearest neighbor interpolation and matches the frame rate of the synchronization features \(\featsync\) with that of the audio latent \(x\).
This frame-aligned conditioning \(c_f\) is injected via the adaLN layers in the audio stream for feature modulation.
Similar to \cref{eq:global_condition}, we apply \(c_f\) via
\begin{equation}
    \text{adaLN}_f(x, c_f) = \text{LayerNorm}(x) \cdot \mathbf{A}_\gamma(c_f) + \mathbf{A}_\beta(c_f), 
\end{equation}
where \(\mathbf{A}_\gamma, \mathbf{A}_\beta\in\mathbb{R}^{h\times h}\) are MLPs.
Different from \cref{eq:global_condition}, the scales and biases are applied per token without broadcasting, providing fine-grained control. %

\subsection{Training and Inference}

\subsubsection{Multimodal Datasets}

\paragraph{VGGSound.}
We train on VGGSound~\cite{chen2020vggsound} as the only audio-text-visual dataset. It offers around 500 hours of footage. 
Additionally, VGGSound contains a class label (310 classes in total) for each video and we use the class names as input following ReWaS~\cite{jeong2024read} and VATT~\cite{liu2024tellhearvideo}. %
We set aside 2K videos from the training set for validation, resulting in a training set of around 180K 10s videos.
We use the first 8s of each video for training.

\paragraph{Audio-text datasets.}
We use AudioCaps~\cite{audiocaps} ({\small\(\sim\)}128 hours, manually captioned), Clotho~\cite{drossos2020clotho} ({\small\(\sim\)}31 hours, manually captioned), and WavCaps~\cite{mei2024wavcaps} ({\small\(\sim\)}7,600 hours, automatically captioned from metadata) as audio-text datasets for training.
Since they do not contain the visual modality, we set all the visual features and synchronization features corresponding to these samples as learnable empty tokens $\varnothing_v$ and $\varnothing_{\mathit{syn}}$ respectively.
For short audios (<16s), we truncate them to 8s for training, as in VGGSound. 
For longer audios, we take up to five non-overlapping crops of 8s each.
This results in a total of 951K audio clip-text pairs.

\paragraph{Overlaps.}
We notice a minor (<1\% of the test sets) train/test data contamination among these datasets. 
For a fair comparison, we have removed the test sets of VGGSound and AudioCaps %
from all training data.
We provide more details in~\Cref{app:sec:overlap}.

\subsubsection{Implementation Details}

\paragraph{Model variants.}
Our default model generates 16kHz audio encoded as 20-dimensional, 31.25fps latents (following Frieren~\cite{wang2024frieren}), with $N_1=4, N_2=8, h=448$.
We refer to this default model as `S-16kHz'.
We additionally train larger models and models with higher audio sampling rates: `S-44.1kHz', `M-44.1kHz', and `L-44.1kHz', detailed in~\Cref{app:sec:model_variants}.
The parameter counts and running time of these models are summarized in~\Cref{tab:main_results}.
We describe additional implementation details in~\Cref{app:sec:training,app:sec:latents,app:sec:networks}.

\paragraph{Classifier-free guidance.}
To enable classifier-free guidance~\cite{ho2022classifier} during inference, we randomly mask away the visual tokens (\(\featvis{}\) and \(\featsync{}\)) or the text with a \(10\%\) probability during training.
The masked visual tokens are replaced with learnable tokens (\(\varnothing_v\) and \(\varnothing_{\mathit{syn}}\)), while any masked text is replaced with the empty string \(\varnothing_t\).

\paragraph{Inference.}
By default, we use Euler's method for numerical integration with 25 steps, with a classifier-free guidance strength of $4.5$.
Both video and text conditions are optional during test-time -- we replace the missing modalities with empty tokens \(\varnothing_v\), \(\varnothing_{\mathit{syn}}\), or \(\varnothing_t\).
Recall, we deliberately do not use any absolute position encoding and thus can generalize to different durations at test time (\eg, 8s in VGGSound and 10s in AudioCaps in~\Cref{sec:main_results}).

%% file: sec/04-expr.tex
\section{Experiments}
\label{sec:expr}

\input{tab/tab-main-results}

\subsection{Metrics}
We assess the generation quality in four different dimensions: distribution matching, audio quality, semantic alignment, and temporal alignment.

\noindent\textbf{Distribution matching} assesses the similarity in feature distribution between ground-truth audio and generated audio, under some embedding models.
Following common practice~\cite{iashin2021taming,wang2024frieren}, we compute Fréchet Distance (FD) and Kullback–Leibler (KL) distance.
For FD, we adopt PaSST~\cite{koutini2021efficient} (\fdpasst), PANNs~\cite{kong2020panns} (\fdpann), and VGGish~\cite{gemmeke2017audio} (\fdvgg)  as embedding models.
Note, PaSST operates at 32kHz, while both PANNs and VGGish operate at 16kHz.
Moreover, both PaSST and PANNs produce global features, while VGGish processes non-overlapping 0.96s clips.
For the KL distance, we adopt PANNs (\klpann{}) and PaSST (\klpasst{}) as classifiers.
We follow the implementation of~\citet{liu2023audioldm}.

\noindent\textbf{Audio quality} assesses the generation quality without comparing it to the ground truth using the Inception Score~\cite{salimans2016improved}. 
We adopt PANNs as the classifier following~\citet{wang2024frieren}.

\noindent\textbf{Semantic alignment} assesses the semantic similarity between the input video and the generated audio.
We use ImageBind~\cite{girdhar2023imagebind} following~\citet{viertola2024temporally} to extract visual features from the input video and audio features from the generated audio and compute the average cosine similarity as ``IB-score''.

\noindent\textbf{Temporal alignment} assesses audio-visual synchrony with a synchronization score (DeSync).
DeSync is predicted by Synchformer~\cite{iashin2024synchformer} as the misalignment (in seconds) between the audio and video.
Note that~\citet{viertola2024temporally} also use the synchronization score but evaluate on audio (2.56s) that is shorter than the context window (4.8s) of Synchformer.
Instead, we evaluate on longer (8s) audios by taking two crops (first 4.8s and last 4.8s) and averaging the results.
Thus, the scores from~\citet{viertola2024temporally} are not directly comparable with ours.

\subsection{Main Results}
\label{sec:main_results}

\begin{figure*}
    \centering
    \includegraphics[width=0.85\linewidth]{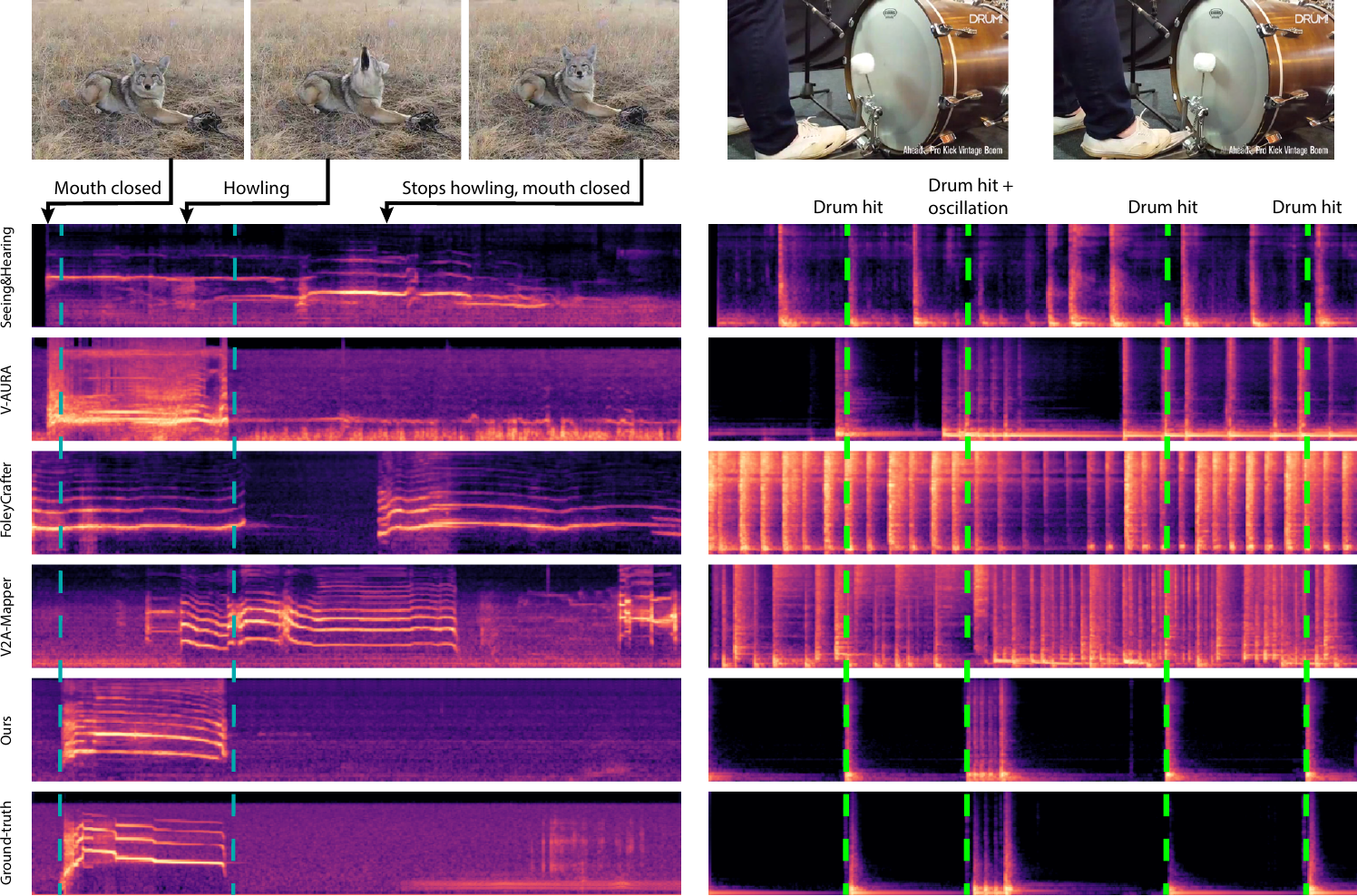}
    \caption{We visualize the spectrograms of generated audio (by prior works and our method) and the ground-truth.
    Note our method generates the audio effects most closely aligned to the ground-truth, while other methods often generate sounds not explained by the visual input and not present in the ground-truth.
    }
    \label{fig:visualization}
\end{figure*}

\paragraph{Video-to-audio.}
\Cref{tab:main_results} compares our main results on the VGGSound~\cite{chen2020vggsound} test set (\({\sim}\)15K videos) with existing state-of-the-art models.
We evaluate all generations at 8s following~\citet{wang2024frieren} by truncating longer audio to 8 seconds. 
For V-AURA~\cite{viertola2024temporally}, we use the official autoregression code to generate 8s audio.
ReWaS~\cite{jeong2024read} only generates 5s audio thus we evaluate it as-is, by truncating the ground truth also to 5s -- we indicate this discrepancy via the gray font in the table.
Our smallest model (157M) demonstrates better distribution matching, audio quality, semantic alignment, and temporal alignment than prior methods, while being fast. A notable exception is the IB-score comparison with Seeing-and-Hearing~\cite{xing2024seeing} (SAH).
We note that SAH directly optimizes the IB score during test time, which we do not perform.
Further, our larger models continue to improve in \fdpasst{} and IB-score, though we observe diminishing returns potentially limited by data quality and the amount of audio-visual data.
Note, while our method uses more data for \emph{multimodal joint training}, we \emph{do not use more data overall} than some of the existing methods: 
FoleyCrafter~\cite{zhang2024foleycrafter}, V2A-Mapper~\cite{wang2024v2a}, ReWaS~\cite{jeong2024read}, and SAH~\cite{xing2024seeing} all finetune/incorporate a text-to-audio model that has been trained on audio-text data similar to the one we use.
For a fair evaluation, we use the precomputed samples provided by~\cite{wang2024frieren,wang2024v2a,liu2024tellhearvideo}, and reproduce the results using the official inference code for~\cite{viertola2024temporally,zhang2024foleycrafter,jeong2024read,xing2024seeing}.
\Cref{fig:visualization} visualizes our results and compares them with prior works.
We present the results of a user study and comparison with Movie Gen Audio~\cite{polyak2024movie} in~\Cref{sec:user_study,sec:movie_gen}.
To address any potential bias introduced by using the model-based DeSync metric, we additionally assess synchronization via model-free metrics on Greatest Hits~\cite{owens2016visually} in \Cref{tab:greatesthits}, detailed in \Cref{sec:app:greatesthits}.

\begin{table}[ht]
\small
    \centering
    \begin{NiceTabular}{l@{\hspace{8pt}}c@{\hspace{8pt}}c@{\hspace{8pt}}c@{\hspace{8pt}}c}
    \toprule
    Method & Acc. \(\uparrow\) & AP \(\uparrow\) & F1\(\uparrow\) & DeSync\(\downarrow\)  \\
    \midrule
    Frieren~\cite{wang2024frieren} & 0.6949 & 0.7846 & 0.6550 & 0.851  \\
    V-AURA~\cite{viertola2024temporally} & 0.5852 & 0.8567 & 0.6441 & 0.654  \\
    FoleyCrafter~\cite{zhang2024foleycrafter} & 0.4533 & 0.6939 & 0.4319 & 1.225  \\
    Seeing\&Hearing~\cite{xing2024seeing} & 0.1156 & 0.8342 & 0.1591 & 1.204  \\
    \rowcolor{defaultColor}
    MMAudio-S-16kHz & \textbf{0.7637} & 0.9010 & \textbf{0.7928} & 0.483  \\
    \rowcolor{defaultColor}
    MMAudio-S-44.1kHz & 0.7150 & \textbf{0.9097} & 0.7666 & 0.444  \\
    \rowcolor{defaultColor}
    MMAudio-M-44.1kHz & 0.7226 & 0.9054 & 0.7620 & 0.443  \\
    \rowcolor{defaultColor}
    MMAudio-L-44.1kHz & 0.7158 & 0.9064 & 0.7535 & \textbf{0.442}  \\
    \midrule
    \bottomrule
    \end{NiceTabular}
    \caption{
    Onset accuracy, average precision (AP), and F1-score on Greatest Hits, with DeSync on VGGSound for reference.
    }
    \label{tab:greatesthits}
\end{table}

\begin{table*}[t]
\begin{minipage}[t]{0.53\linewidth}
\input{tab/tab-text-to-audio} 
\end{minipage}
\quad
\begin{minipage}[t]{0.45\linewidth}
\input{tab/tab-cross-modal}

\end{minipage}
\end{table*}

\paragraph{Text-to-audio.}
Our multimodal framework can be applied to \emph{text-to-audio} synthesis without fine-tuning. 
\Cref{tab:text_to_audio} compares our method with state-of-the-art text-to-audio models using the AudioCaps~\cite{audiocaps} test set.
For a fair comparison, we follow the evaluation protocol of GenAU~\cite{haji2024taming} to evaluate 10s samples in the AudioCaps~\cite{audiocaps} test set without using CLAP re-ranking (used by AudioLDM~\cite{liu2023audioldm}).
We transcribe the baselines directly from~\citet{haji2024taming}, who have reproduced those results using officially released checkpoints under the same evaluation protocol.
We assess \fdpann{}, \fdvgg{}, IS, and CLAP~\cite{wu2023large}. 
CLAP measures the semantic alignment between the generated audio and the input caption.
While our main focus is on video-to-audio synthesis, MMAudio attains state-of-the-art semantic alignment (CLAP) and audio quality (IS), due to a rich semantic feature space learned from multimodal joint training.
We note that we attain a worse \fdvgg{} score compared to recent works.
We hypothesize that this is because VGGish processes local features (clips of 0.96s) while our strength lies in generating globally and semantically consistent audio.

\subsection{Ablations}
We base all ablations on the small-16kHz model and evaluate distribution matching (\fdpasst), audio quality (IS), semantic alignment (IB-score), and temporal alignment (DeSync) on the VGGSound~\cite{chen2020vggsound} test set.
We highlight our default setting using a  \colorbox{defaultColor}{blue} background.

\paragraph{Cross-modal alignment.}
To elucidate the benefits of joint multimodal training, we mask away some modalities during training and observe the effects on the results, summarized in~\Cref{tab:cross-modal}.
We denote the setting as (modalities used for audio-visual-text data + modalities used for audio-text data), where A: Audio, V: Video, T: Text.
Our default setting (AVT+AT) means that we train on audio-visual-text data (VGGSound class labels as text input) and audio-text data.
We make three observations:
\begin{enumerate}
    \item Masking away the text modality from either the former (AV+AT) or the latter (AVT+A) leads to worse results. 
    This suggests that having a joint ``text feature space'' is beneficial for multimodal training.
    \item Adding uncaptioned audio data improves results (AVT \vs AVT+A). 
    This suggests that our network even benefits from training on unconditional generation data by learning the distribution of natural sounds.
    \item When no audio-text data is used, using the simple class labels in VGGSound does not affect results significantly (AVT \vs AV).
    This suggests that training on large multimodal datasets, rather than adding a transformer branch in the network or using the class labels, is key.
\end{enumerate}

\input{tab/tab-mm-data}

\paragraph{Multimodal data.}
Training on a large collection of multimodal data is crucial. 
\Cref{tab:multimodal_data} shows our model's performance when we vary the amount of audio-text training data.
We always sample audio-visual data and audio-text data at the same (roughly 1:1, see \Cref{app:sec:training}) ratio except when we use no audio-text data. %
When we do not use audio-text data, we observe overfitting and stop training early.
When more multimodal data is used, distribution matching (\fdpasst), semantic alignment (IB-score), and temporal alignment (DeSync) improve with diminishing returns.

\paragraph{Conditional synchronization module.}
We compare several different methods for incorporating synchronization features: 
1) our default of using the conditional synchronization module (\Cref{sec:cond_sync}); 
2) incorporating the synchronization features into the visual branch of the multimodal transformer. 
Concretely, we upsample (with nearest neighbor) the CLIP features to 24fps, and then sum the CLIP features and Sync features after linear projections as the final visual feature.
We illustrate this architecture in~\Cref{fig:sum_sync_with_visual};
finally, 3) not using synchronization features.
\Cref{tab:sync_vary} (top) shows that our synchronization module attains the best temporal alignment.
We note the ``sum sync with visual'' method achieves higher audio quality (IS) -- we hypothesize that since upsampling CLIP features increased the number of tokens in the visual stream by three times, the model benefits from using the longer sequence for more fine-grained computations.

\input{tab/tab-sync}

\paragraph{RoPE embeddings.}
We compare our aligned RoPE formulation with 1) not using RoPE embeddings and 2) non-aligned RoPE embeddings, \ie, no frequency scaling in the visual branch.
\Cref{tab:sync_vary} (bottom) shows that using aligned RoPE embeddings~\cite{su2024roformer} improves audio-visual synchrony. 

\input{tab/tab-arch}

\paragraph{ConvMLP.}
\Cref{tab:arch} (top) summarizes the performance difference of using MLP \vs ConvMLP.
For the MLP model, we increase $h$ from $448\to512$, $N_1$ from $4\to6$, and $N_2$ from $8\to10$ to roughly match the number of parameters in the ConvMLP model.
The ConvMLP model is better at capturing local temporal structure and thus has a better performance, especially in synchronization.

\paragraph{Ratio between $N_1$ and $N_2$.}
\Cref{tab:arch} (bottom) compares different assignments of the number of multimodal (\(N_1\)) and single-modal (\(N_2\)) transformer blocks with roughly the same parameter budget.
We note our default assignment (\(N_1=4, N_2=8\)) performs similarly as using more single-modal blocks (\(N_1=2, N_2=13\)) and better than using fewer single-modal blocks (\(N_1=6, N_2=3\)).
We think this is because using single-modal blocks allows us to build a deeper network with the same parameter count.

\subsection{Limitations}
Our model generates unintelligible mumbles when prompted to generate human speech (from seeing mouth movement or from text input).
We believe human speech is inherently more complex (\eg, with languages, tones, and grammars) and our model aimed at general audio effects (Foley) fails to adequately accommodate.

%% file: tab/tab-main-results.tex
\begin{table*}[t]
\small
    \centering
    \begin{NiceTabular}{l@{\hspace{4pt}}c@{\hspace{4pt}}c@{\hspace{4pt}}c@{\hspace{4pt}}c@{\hspace{4pt}}c@{\hspace{4pt}}c@{\hspace{1pt}}c@{\hspace{1pt}}c@{\hspace{1pt}}c@{\hspace{1pt}}c}
    \toprule
    Method & & \multicolumn{5}{c}{{\footnotesize Distribution matching}} & {\footnotesize Audio quality} & {\footnotesize Semantic align.} & {\footnotesize Temporal align.} & \\
    \cmidrule(lr{\dimexpr 4\tabcolsep-16pt}){3-7}
    \cmidrule(lr{\dimexpr 4\tabcolsep-16pt}){8-8}
    \cmidrule(lr{\dimexpr 4\tabcolsep-16pt}){9-9}
    \cmidrule(lr{\dimexpr 4\tabcolsep-16pt}){10-10}
    & Params & \fdpasst$\downarrow$ & \fdpann$\downarrow$ & \fdvgg$\downarrow$ & \klpann$\downarrow$ & \klpasst$\downarrow$ & \ispann$\uparrow$ & IB-score$\uparrow$ & DeSync$\downarrow$ & Time (s)$\downarrow$ \\
    \midrule
    \textcolor{gray}{ReWaS}~\cite{jeong2024read}$^\ast$ & 619M & \textcolor{gray}{141.38} & \textcolor{gray}{17.54} & \textcolor{gray}{1.79} & \textcolor{gray}{2.87} & \textcolor{gray}{2.82} & \textcolor{gray}{8.51} & \textcolor{gray}{14.82} & \textcolor{gray}{1.062} & \textcolor{gray}{15.97} \\
    Seeing\&Hearing~\cite{xing2024seeing}$^\ast$ & 415M & 219.01 & 24.58 & 5.40 & 2.26 & 2.30 & 8.58 & \textbf{33.99} & 1.204 & 14.55 \\
    V-AURA~\cite{viertola2024temporally}$^{\ast\lozenge}$ & 695M  & 218.50 & 14.80 & 2.88 & 2.42 & 2.07 & 10.08 & 27.64 & 0.654 & 16.55 \\
    VATT~\cite{liu2024tellhearvideo}$^\dagger$ & - & 131.88 & 10.63 & 2.77 & \textbf{1.48} & 1.41 & 11.90 & 25.00 & 1.195 & - \\
    Frieren~\cite{wang2024frieren}$^{\dagger\lozenge}$ & 159M  & 106.10 & 11.45 & 1.34 & 2.73 & 2.86 & 12.25 & 22.78 & 0.851 & - \\
    FoleyCrafter~\cite{zhang2024foleycrafter}$^\ast$ & 1.22B & 140.09 & 16.24 & 2.51 & 2.30 & 2.23 & 15.68 & 25.68 & 1.225 & 1.67 \\
    V2A-Mapper~\cite{wang2024v2a}$^{\dagger\lozenge}$ & 229M & 84.57 & 8.40 & 0.84 & 2.69 & 2.56 & 12.47 & 22.58 & 1.225 & - \\
    \rowcolor{defaultColor}
    MMAudio-S-16kHz & 157M & 70.19  & 5.22 & \textbf{0.79} & 1.65 & 1.59 & 14.44 & 29.13 & 0.483 & \textbf{1.23} \\
    \rowcolor{defaultColor}
    MMAudio-S-44.1kHz & 157M  & 65.25 & 5.55& 1.66 & 1.67 & 1.44 & \textbf{18.02} & 32.27 & 0.444 & 1.30\\
    \rowcolor{defaultColor}
    MMAudio-M-44.1kHz  & 621M  & 61.88 & 4.74 & 1.13 & 1.66 & 1.41 & 17.41 & 32.99 & 0.443 & 1.35 \\
    \rowcolor{defaultColor}
    MMAudio-L-44.1kHz & 1.03B & \textbf{60.60} & \textbf{4.72} & 0.97 & 1.65 & \textbf{1.40} & 17.40 & 33.22 & \textbf{0.442} & 1.96 \\
    \midrule
    \bottomrule
    \end{NiceTabular}
    \caption{Video-to-audio results on the VGGSound test set.
    Following the common practice~\cite{wang2024frieren}, the parameter counts exclude pretrained feature extractors (\eg, CLIP), latent space encoders/decoders, and vocoders.
    Time is the total running time using the official code to generate one sample with a batch size of one after warm-up and excludes any disk I/O operations on an H100 GPU.
    $\ast$: reproduced using official evaluation code. 
    $\dagger$: evaluated using generation samples obtained directly from the authors.
    $\lozenge$: does not use text input during testing.
    Note, Seeing\&Hearing~\cite{xing2024seeing} directly optimizes ImageBind score during test time, therefore attains the highest IB-score.
    }
    \label{tab:main_results}
\end{table*}

%% file: tab/tab-text-to-audio.tex
\small
    \centering
    \begin{NiceTabular}{l@{\hspace{4pt}}c@{\hspace{6pt}}c@{\hspace{6pt}}c@{\hspace{6pt}}c@{\hspace{6pt}}c}
    \toprule
    Method & Params & \fdpann\(\downarrow\) & \fdvgg\(\downarrow\) & IS\(\uparrow\) & CLAP\(\uparrow\) \\
    \midrule
    AudioLDM 2-L~\cite{liu2024audioldm2} & 712M & 32.50 & 5.11 & 8.54 & 0.212 \\
    TANGO~\cite{ghosal2023tango} & 866M & 26.13 & 1.87 & 8.23 & 0.185 \\
    TANGO 2~\cite{majumder2024tango2} & 866M & 19.77 & 2.74 & 8.45 & 0.264 \\
    Make-An-Audio~\cite{huang2023make} & 453M & 27.93 & 2.59 & 7.44 & 0.207 \\
    Make-An-Audio 2~\cite{huang2023make2} & 937M & 15.34 & 1.27 & 9.58 & 0.251 \\
    GenAU-Large~\cite{haji2024taming} & 1.25B & 16.51 & \textbf{1.21} & 11.75 & 0.285 \\
    \rowcolor{defaultColor}
    MMAudio-S-16kHz & 157M & 14.42 & 2.98 & 11.36 & 0.282 \\
    \rowcolor{defaultColor}
    MMAudio-S-44.1kHz & 157M & 15.26 & 2.74 & 11.32 & 0.331 \\
    \rowcolor{defaultColor}
    MMAudio-M-44.1kHz & 621M & \textbf{14.38} & 4.07 & 12.02 & \textbf{0.351} \\
    \rowcolor{defaultColor}
    MMAudio-L-44.1kHz & 1.03B & 15.04 & 4.03 & \textbf{12.08} & 0.348 \\
    \midrule
    \bottomrule
    \end{NiceTabular}
    \caption{Text-to-audio results on the AudioCaps test set.
    For a fair comparison, we follow the evaluation protocol of~\cite{haji2024taming} and transcribe all baselines directly from~\cite{haji2024taming}, who have reproduced those results using officially released checkpoints under the same evaluation protocol.
    }
    \label{tab:text_to_audio}

%% file: tab/tab-cross-modal.tex
\small
    \centering
    \begin{NiceTabular}{l@{\hspace{4pt}}c@{\hspace{4pt}}c@{\hspace{4pt}}c@{\hspace{4pt}}c}
    \toprule
    Training modalities & \fdpasst\(\downarrow\) & IS\(\uparrow\) & IB-score\(\uparrow\) & DeSync\(\downarrow\) \\
    \midrule
    \rowcolor{defaultColor}
    AVT+AT & \textbf{70.19} & \textbf{14.44} & \textbf{29.13} & \textbf{0.483} \\
    AV+AT & 72.77 & 12.88 & 28.10 & 0.502 \\
    AVT+A & 71.01 & 14.30 & 28.72 & 0.496 \\
    AVT & 77.38 & 12.53 & 27.98 & 0.562 \\
    AV & 77.27 & 12.69 & 28.10 & 0.502 \\
    \midrule
    \bottomrule
    \end{NiceTabular}
    \caption{Results when we vary the training modalities. 
    A:~Audio, V:~Video, T:~Text.
    In the second and third rows, we mask away the text token in either audio-visual data or audio-text data.
    In the last two rows, we do not use any audio-text data.
    }
    \label{tab:cross-modal}

%% file: tab/tab-mm-data.tex
\begin{table}
\small
    \centering
    \begin{NiceTabular}{l@{\hspace{8pt}}c@{\hspace{8pt}}c@{\hspace{8pt}}c@{\hspace{8pt}}c}
    \toprule
    \% audio-text data & \fdpasst\(\downarrow\) & IS\(\uparrow\) & IB-score\(\uparrow\)  & DeSync\(\downarrow\) \\
    \midrule
    \rowcolor{defaultColor}
    100\% & \textbf{70.19} & 14.44 & \textbf{29.13} & \textbf{0.483} \\
    50\% & 71.03  & \textbf{14.62} & 29.11  & 0.489 \\
    25\% & 71.67 & 14.41 & 28.75  & 0.505 \\
    10\% & 79.21 & 13.55 & 27.47  & 0.514 \\
    \midrule
    None & 77.38 & 12.53  & 27.98 & 0.562 \\
    \midrule
    \bottomrule
    \end{NiceTabular}
    \caption{Results when we vary the amount of multimodal training data.
    For the first four rows, we sample audio-visual and audio-text data at a 1:1 ratio during training.
    For the last row, only audio-visual data is used.}
    \label{tab:multimodal_data}
\end{table}

%% file: tab/tab-sync.tex
\begin{table}
\small
    \centering
    \begin{NiceTabular}{l@{\hspace{6pt}}c@{\hspace{6pt}}c@{\hspace{6pt}}c@{\hspace{6pt}}c}
    \toprule
    Variant & \fdpasst\(\downarrow\) & IS\(\uparrow\)  & IB-score\(\uparrow\) & DeSync\(\downarrow\) \\
    \midrule
    \rowcolor{defaultColor}
    With sync module & 70.19 & 14.44 & 29.13 & \textbf{0.483} \\
    Sum sync with visual & 73.59 & \textbf{16.70} & 28.65 & 0.490 \\
    No sync features & \textbf{69.33} & 15.05 & \textbf{29.31} & 0.973 \\
    \midrule
    \rowcolor{defaultColor}
    Aligned RoPE & \textbf{70.19} & 14.44 & 29.13 & \textbf{0.483} \\
    No RoPE & 70.24 & \textbf{14.54} & 29.23 & 0.509 \\
    Non-aligned RoPE & 70.25 & \textbf{14.54} & \textbf{29.25} & 0.496 \\
    \midrule
    \bottomrule
    \end{NiceTabular}
    \caption{Results when we use synchronization features or RoPE embeddings differently.}
    \label{tab:sync_vary}
\end{table}

%% file: tab/tab-arch.tex
\begin{table}
\small
    \centering
    \begin{NiceTabular}{l@{\hspace{8pt}}c@{\hspace{8pt}}c@{\hspace{8pt}}c@{\hspace{8pt}}c}
    \toprule
    Variant & \fdpasst\(\downarrow\) & IS\(\uparrow\) & IB-score\(\uparrow\) & DeSync\(\downarrow\) \\
    \midrule
    \rowcolor{defaultColor}
    ConvMLP & \textbf{70.19 } & \textbf{14.44} & \textbf{29.13} & \textbf{0.483} \\
    MLP & 73.84 & 13.01 & 28.99 & 0.533 \\
    \midrule
    \rowcolor{defaultColor}
    \(N_1=4, N_2=8\) & \textbf{70.19} & 14.44 & 29.13 & \textbf{0.483} \\
    \(N_1=2, N_2=13\) & 70.33 & \textbf{15.18} & \textbf{29.39} & 0.487 \\
    \(N_1=6, N_2=3\) & 72.53 & 13.75 & 29.06 & 0.509 \\
    \midrule
    \bottomrule
    \end{NiceTabular}
    \caption{Results when we vary the MLP architecture or the ratio between multi-/single-modality transformer blocks.}
    \label{tab:arch}
\end{table}

%% file: sec/05-conclusion.tex
\section{Conclusion}
\label{sec:conclusion}

We propose MMAudio, the first multimodal training pipeline that jointly considers audio, video, and text modalities, resulting in effective data scaling and cross-modal semantic alignment.
Combined with a conditional synchronization module, our method achieves a new state-of-the-art performance among public models and comparable performance with Movie Gen Audio.
We believe a multimodal formulation is key for the synthesis of data in any modality and MMAudio lays the foundation in the audio-video-text space. 

\clearpage

\paragraph{Acknowledgment.}
This work is supported in part by Sony. 
AS is supported by NSF grants 2008387, 2045586, 2106825, and NIFA award 2020-67021-32799.
We sincerely thank Kazuki Shimada and Zhi Zhong for their helpful feedback on this manuscript.

%% file: sec/10-appendix.tex
\beginsupplement

\titlespacing*{\section}{0pt}{3.5ex plus 1ex minus .2ex}{2.3ex plus .2ex}
\titlespacing*{\subsection}{0pt}{3.25ex plus 1ex minus .2ex}{1.5ex plus .2ex}
\titlespacing*{\subsubsection}{0pt}{3.25ex plus 1ex minus .2ex}{1.5ex plus .2ex}
\titlespacing*{\paragraph}{0em}{.5ex plus .5ex minus .3ex}{1em}

\renewcommand{\contentsname}{Table of Contents}
{
\large
\tableofcontents
}

\clearpage

\section{User Study}
\label{sec:user_study}
In addition to the objective metrics presented in~\Cref{tab:main_results}, we have also performed a user study for subjective evaluation on the VGGSound~\cite{chen2020vggsound} test set.
For comparisons, we have selected our best model (MMAudio-L-44.1kHz) and four best baselines:

\begin{enumerate}[\hspace{0.5em}1.]
    \item Seeing and Hearing~\cite{xing2024seeing}, as it has the highest ImageBind (\ie, best semantic alignment with videos) score, besides ours.
    \item V-AURA~\cite{viertola2024temporally}, as it has the lowest DeSync (\ie, best temporal alignment) with videos, besides ours.
    \item VATT~\cite{liu2024tellhearvideo}, as it has the lowest Kullback–Leibler divergence (\ie, \klpann{} and \klpasst{}), besides ours.
    \item V2A-Mapper~\cite{wang2024v2a}, as it has the lowest Fréchet distances (\ie, \fdpasst{}, \fdpann{}, and \fdvgg{}), besides ours.
\end{enumerate}

We sample eight videos from the VGGSound~\cite{chen2020vggsound} test set, after excluding videos that are of low-resolution (below 360p) or that contain human speech.
In total, each participant evaluates 40 videos (8 videos \(\times\) 5 methods).
We group the samples for the same video, and randomly shuffle the ordering in each group to avoid bias.
We ask each participant to rate the generation in three aspects using the Likert scale~\cite{likert1932technique} (1-5; strongly disagree, disagree, neutral, agree, strongly agree) providing the following instructions:

\begin{enumerate}[\hspace{0.5em}(a)]
    \item The audio is of \textbf{high quality}.
    
    \emph{Explanation}: An audio is low-quality if it is noisy, unclear, or muffled. In this aspect, ignore visual information and focus on the audio.

    \item The audio is \textbf{semantically aligned} with the video.

    \emph{Explanation}: An audio is semantically misaligned with the video if the audio effects are unlikely to occur in the scenario depicted by the video, \eg, the sound of an explosion in a library.

    \item The audio is \textbf{temporally aligned} with the video.
    
    \emph{Explanation}: An audio is temporally misaligned with the video if the audio sounds delayed/advanced compared to the video, or when audio events happen at the wrong time (\eg, in the video, the drummer hits the drum twice and stops; but in the audio, the sound of the drum keeps occurring).
\end{enumerate}

In total, we have collected 920 responses in each of these aspects from 23 participants.
\Cref{tab:user_study} summarizes the results from the user study.
MMAudio receives significantly higher ratings in all three aspects from the users, which aligns with the objective metrics presented in~\Cref{tab:main_results} of the main paper.

\input{tab/tab-user-study}

\section{Comparisons with Movie Gen Audio}
\label{sec:movie_gen}

Recently, Movie Gen Audio~\cite{polyak2024movie} has been introduced for generating sound effects and music for input videos.
While Movie Gen Audio's technical details are sparse, it represents the industry's current state-of-the-art video-to-audio synthesis algorithm.
Its 13-billion parameters model has been trained on non-publicly accessible data that is \(>100\times\) larger than ours.
Nevertheless, we compare MMAudio to Movie Gen Audio~\cite{polyak2024movie} to benchmark the differences between public and private models.

At the time of writing, the only accessible outputs from Movie Gen Audio are 527\footnote{While the MovieGen technical report mentioned 538 samples, only 527 were released at the time of writing.}
generations in the ``Movie Gen Audio Bench'' dataset.
All the videos from Movie Gen Audio Bench are generated by MovieGen~\cite{polyak2024movie}, which we note is different from the distribution of real-world videos (\eg, over-smoothed textures, slow motions).
Since these are synthetic videos, there is no corresponding ground-truth audio. 
We run our best model MMAudio-L-44.1kHz on these videos and the corresponding audio prompts (which Movie Gen Audio also uses) and compare our generations with Movie Gen Audio.

Since there is no ground truth audio, among the standard metrics that we have used in the main paper, we can only evaluate Inception Score (IS, audio quality), IB-score (ImageBind~\cite{girdhar2023imagebind} similarly, semantic alignment between video and audio), DeSync (misalignment predicted by SynchFormer~\cite{iashin2024synchformer} between video and audio), and CLAP~\cite{wu2023large,elizalde2024natural} (alignment between text and audio).
Additionally, we have conducted a user study following the protocol of \Cref{sec:user_study}, and have excluded audios with very low volume (cannot be heard clearly at a normal volume) generated by Movie Gen Audio to prevent bias.
We sampled a total of 5 videos and received 230 responses in each of the aspects from 23 participants.

\Cref{tab:moviegen} summarizes our results. 
In subjective metrics, MMAudio is comparable to Movie Gen Audio -- slightly worse in semantic alignment and slightly better in temporal alignment.
In objective metrics, we observe the same trend -- MMAudio and Movie Gen Audio obtain the same audio quality (IS) score, Movie Gen Audio has a better semantic alignment (IB-score and CLAP), and MMAudio has a better video-audio synchrony (DeSync).

\input{tab/tab-moviegen}

\begin{figure}[ht]
\vspace{1ex}
    \centering
    \includegraphics[width=\linewidth]{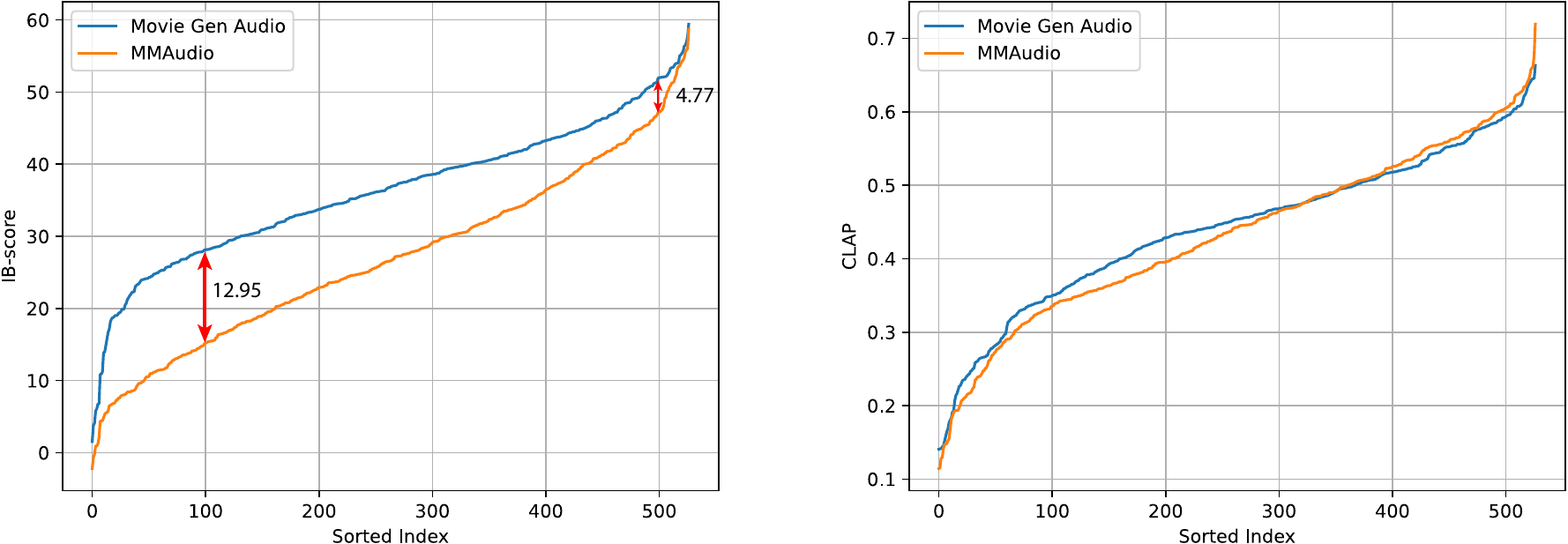}
    \caption{
    Sorted MMAudio and Movie Gen Audio performance scores in Movie Gen Audio Bench.
    }
    \label{fig:moviegen_plots}
\end{figure}

\begin{figure}[ht]
\small
    \centering
     \begin{minipage}{0.48\textwidth}
        \includegraphics[width=\linewidth]{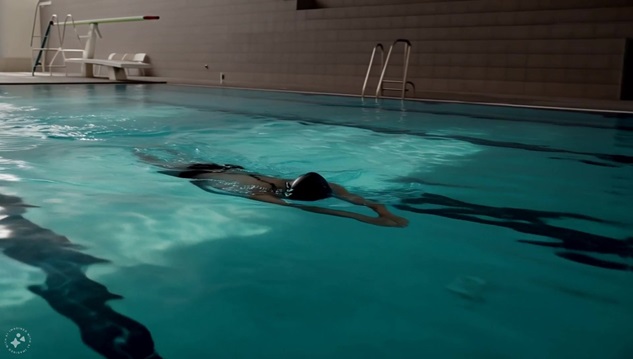}
        Audio prompt: rhythmic splashing and lapping of water

        IB-score (Movie Gen Audio): 42.74

        IB-score (MMAudio, ours): 53.95
    \end{minipage}
    \begin{minipage}{0.48\textwidth}
        \includegraphics[width=\linewidth]{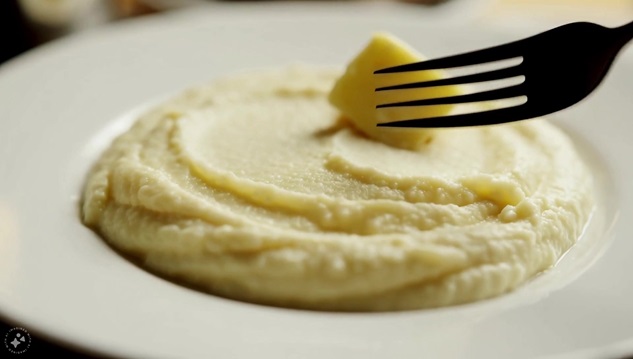}
        Audio prompt: creamy sound of mashed potatoes being scooped

        IB-score (Movie Gen Audio): 30.94

        IB-score (MMAudio, ours): 10.52
    \end{minipage}
    \vspace{2ex}
    \caption{
    Examples of videos in Movie Gen Audio Bench that are well/not well covered by our training data.
    Left: with a familiar concept in our training data (516 swimming videos in the VGGSound training set), MMAudio achieves a higher IB-score.
    Right: with an unfamiliar concept (there are no videos about mashed potatoes in VGGSound~\cite{chen2020vggsound}, according to the provided labels), MMAudio attains a significantly lower IB-score.
    }
    \label{fig:moviegen_examples}
\end{figure}

Further, in terms of IB-score, we find that MMAudio struggles more in some videos, while Movie Gen Audio delivers more consistent results. 
We plot the sorted IB-score comparing MMAudio and Movie Gen Audio in \Cref{fig:moviegen_plots} (left).
Movie Gen Audio consistently performs better in the low-performance regime, but the gap narrows in the high-performance region.
We believe this is due to our limited training data, which is unable to adequately cover the data in Movie Gen Audio Bench and thus falls short in unfamiliar video types.
Note, our only video-audio dataset for training is VGGSound~\cite{chen2020vggsound} which contains videos for 310 classes.
We hypothesize that collecting open-world data beyond these classes can effectively reduce this performance gap.
The same phenomenon occurs at a much smaller scale for the CLAP score, which might be because we use more audio-text data.~\Cref{fig:moviegen_examples} shows examples where we obtain a substantially higher/lower IB-score on videos with concepts well/not well covered by the training data.

\section{Evaluation on the Greatest Hits Dataset}
\label{sec:app:greatesthits}

To address any potential bias by using the model-based DeSync metric, we conduct an additional experiment to assess temporal alignment by comparing the onsets of generated audio with ground-truth labels.
Concretely, we use the \emph{Greatest Hits}~\cite{owens2016visually} test set (244 videos) which contains videos with distinct and labeled sound events (\emph{a drumstick hitting <object>}). 
Notably, neither our models nor the baselines have been trained on this dataset. 
We test both our method and baselines (using available code) on each video's first 8 seconds (due to the models' constraints):  
we extract onsets from the generated audio following~\cite{du2023conditional} and compare them with the labeled sound events.
We assess performance using accuracy, average precision (AP), and F1-score.
We provide the results in \Cref{tab:greatesthits} and visualize the spectrograms in \Cref{fig:spectrogram_vis_greatesthits}.
MMAudio achieves significantly better performance in these \emph{model-free} metrics.
Note that a high AP (not accuracy or F1) can be achieved by generating very few onsets (\eg, silent/noise), which is the case in Seeing\&Hearing.

\section{Ablations on Filling in Missing Modalities}

Among our training data, VGGSound is the only tri-modal (with class names as text) dataset while all others are audio-text.
For other data, we replace missing visual modalities (CLIP and Sync features) with end-to-end learnable embeddings (\(\varnothing_{\mathit{v}}\) and \(\varnothing_{\mathit{syn}}\)) and missing text modalities with the empty string (\(\varnothing_{\mathit{t}}\)).
We believe other methods to fill in missing modalities would be similarly effective since the deep net likely adapts.
Indeed, replacing the missing modalities with either all learnable embeddings or zeros yields no significant difference (\Cref{tab:masking}).
Note, we also drop modalities randomly during training to enable classifier-free guidance, which enhances the model's robustness to missing modalities.

\begin{table}[ht]
\small
    \centering
    \begin{NiceTabular}{l@{\hspace{8pt}}c@{\hspace{8pt}}c@{\hspace{8pt}}c@{\hspace{8pt}}c}
    \toprule
    Method & \fdpasst\(\downarrow\) & IS\(\uparrow\) & IB-score\(\uparrow\)  & DeSync\(\downarrow\) \\
    \midrule
    \rowcolor{defaultColor}
    Ours & 70.19 & 14.44 & 29.13 & 0.483 \\
    With all learnable & 70.13 & 14.63 & 29.23  & 0.494 \\
    With zeros & 69.91 & 14.60 & 29.22  & 0.496 \\
    \midrule
    \bottomrule
    \end{NiceTabular}
    \caption{Comparisons of different methods to fill in missing modalities.
    As expected, there is no significant difference as the deep net  learns to adapt.
    }
    \label{tab:masking}
\end{table}

\section{Details on Data Overlaps}\label{app:sec:overlap}
We note that there are training and testing data overlaps among commonly used datasets for video-to-audio generation.
For example, AudioSet~\cite{gemmeke2017audio} is commonly used to train VAE encoders/decoders but it contains test set data from VGGSound~\cite{chen2020vggsound} and AudioCaps~\cite{audiocaps}.
Additionally, AudioCaps is often used to train text-to-audio models~\cite{xue2024auffusion}, which is then used as the backbone for video-to-audio models which evaluate on VGGSound~\cite{chen2020vggsound} -- however, part of the VGGSound test set overlaps with the AudioCaps training set. 
Moreover, AVSync15~\cite{zhang2024audio}, which is sometimes used jointly with VGGSound for training/evaluating video-to-audio algorithms~\cite{zhang2024foleycrafter}, contains severe cross-contamination with VGGSound. 
This results in biased evaluations in both VGGSound and AVSync15.
To our best knowledge, this data contamination is not yet addressed in the video-to-audio community. 
We thank \citet{labb2024conette} for raising this issue in the audio captioning field, which has helped us identify this problem.

\Cref{tab:data_overlap} summarizes the observed overlaps.
The overlaps with WavCaps~\cite{mei2024wavcaps} and Freesound~\cite{laion_audio_dataset} have been included as part of their release, which we do not repeat in our table.

We have carefully removed from our training data (AudioSet~\cite{gemmeke2017audio}, AudioCaps~\cite{audiocaps}, Clotho~\cite{drossos2020clotho}, Freesound~\cite{laion_audio_dataset}, WavCaps~\cite{mei2024wavcaps}, and VGGSound~\cite{chen2020vggsound}) anything that overlaps with any of the test sets (VGGSound and AudioCaps).
Additionally, we have also removed from our training data the test set of Clotho~\cite{drossos2020clotho}. Since most baselines have  been trained on VGGSound, we elect not to evaluate on AVSync15.

\input{tab/tab-overlap}

\section{Details on the Audio Latents}\label{app:sec:latents}

As mentioned in the main paper, we obtain the audio latents by first transforming audio waveforms with the short-time Fourier transform (STFT) and extracting the magnitude component as mel spectrograms~\cite{stevens1937scale}.
Then, spectrograms are  encoded into latents by a pretrained variational autoencoder (VAE)~\cite{vae}.
During testing, the generated latents are decoded by the VAE into spectrograms, which are then vocoded by a pretrained vocoder~\cite{lee2022bigvgan} into audio waveforms.
\Cref{tab:stft_settings} tabulates our STFT parameters and latent information.

For the VAE, we follow the 1D convolutional network design of Make-An-Audio~2~\cite{huang2023make2} with a downsampling factor of 2 and trained with reconstruction, adversarial, and Kullback–Leibler divergence (KL) objectives.
We note that the default setting leads to extreme values in the latent at the end of every sequence (\(\pm10\sigma\) away).
To tackle this problem, we have applied the magnitude-preserving network design from EDM2~\cite{karras2024analyzing}, by replacing the convolutional, normalization, addition, and concatenation layers with magnitude-preserving equivalents.
While this change removes the extreme values, it leads to no significant empirical performance difference.
We train the 16kHz model on AudioSet~\cite{gemmeke2017audio}, following Make-An-Audio~2~\cite{huang2023make2}.
For the 44.1kHz model, we increase the hidden dimension from 384 to 512 and train it on AudioSet~\cite{gemmeke2017audio} and Freesound~\cite{laion_audio_dataset} to accommodate the increased reconstruction difficulty due to a higher sampling rate.

For vocoders, we use the BigVGAN~\cite{lee2022bigvgan} trained by Make-An-Audio~2~\cite{huang2023make2} in our 16kHz model.
For our 44.1kHz model, we use BigVGAN-v2~\cite{lee2022bigvgan} (the {\tt bigvgan\_v2\_44khz\_128band\_512x} checkpoint).

\begin{table}[ht]
\small
\vspace{1ex}
    \centering
    \begin{tabular}{lccccccc}
    \toprule
    Model variants & Latent frame rate & \# latent channels & \# mel bins & \# FFTs & Hop size & Window size & Window function \\
    \midrule
    16kHz & 31.25 & 20 & 80 & 1024 & 256 & 1024 & Hann \\
    44.1kHz & 43.07 & 40 & 128 & 2048 & 512 & 2048 & Hann \\
    \midrule
    \bottomrule
    \end{tabular}
    \vspace{1ex}
    \caption{Short-time Fourier transform (STFT) parameters and latent information.}
    \label{tab:stft_settings}
\vspace{1ex}
\end{table}

\section{Network Details}\label{app:sec:networks}

\subsection{Model Variants}\label{app:sec:model_variants}
Our default model generates 16kHz audio encoded as 20-dimensional, 31.25fps latents (following Frieren~\cite{wang2024frieren}), with $N_1=4, N_2=8, h=448$.
We refer to this default model as `S-16kHz'.
To faithfully capture higher frequencies, we also train a 44.1kHz model (`S-44.1kHz') that generates 40-dimensional, 43.07fps latents while all other settings are identical to the default.
To scale up the high-frequency model, we first double the hidden dimension to match the doubled latent dimension, \ie, we use $N_1=4, N_2=8, h=896$ and refer to this model using `M-44.1kHz'.
Finally, we scale the number of layers, \ie, $N_1=7, N_2=14, h=896$ and refer to this model via `L-44.1kHz'.
These model variants are summarized in~\Cref{app:tab:model_variants}.

\begin{table}[ht]
\small
\vspace{1ex}
    \centering
    \begin{tabular}{lcccccc}
    \toprule
    Model variants & Params & \# multimodal blocks \(N_1\) & \# single-modal blocks \(N_2\) & Hidden dim \(h\) & Latent dim & Time (s) \\
    \midrule
    S-16kHz & 157M & 4 & 8 & 448 & 20 & 1.23 \\
    S-44.1kHz & 157M & 4 & 8 & 448 & 40 & 1.30 \\
    M-44.1kHz & 621M & 4 & 8 & 896 & 40 & 1.35 \\
    L-44.1kHz & 1.03B & 7 & 14 & 896 & 40 & 1.96 \\
    \midrule
    \bottomrule
    \end{tabular}
    \vspace{1ex}
    \caption{Summary for different MMAudio model variants.
    Time is the total running time to generate one sample with a batch size of one after warm-up and excludes any disk I/O operations on an H100 GPU.
    }
    \label{app:tab:model_variants}
\vspace{1ex}
\end{table}

\subsection{Projection Layers}

We use projection layers to project input text, visual, and audio features to the hidden dimension \(h\) and for initial aggregation of the temporal context.

\paragraph{Text feature projection.}
We use a linear layer that projects to \(h\), followed by an MLP.

\paragraph{Clip feature projection.}
We use a linear layer that projects to \(h\), followed by a ConvMLP with a kernel size of 3 and a padding of 1.

\paragraph{Sync feature projection.}
We use a 1D convolutional layer with a kernel size of 7 and a padding of 3 that projects to \(h\), an SELU~\cite{klambauer2017self} activation layer, followed by a ConvMLP with a kernel size of 3 and a padding of 1.

\paragraph{Audio feature projection.}
We use a 1D convolutional layer with a kernel size of 7 and a padding of 3 that projects to \(h\), an SELU~\cite{klambauer2017self} activation layer, followed by a ConvMLP with a kernel size of 7 and a padding of 3.

\subsection{Gating}

The gating layers are similar to the adaptive normalization layers (adaLN).
Each global gating layer modulates its input \(y\in\mathbb{R}^{L\times h}\) (\(L\) is the sequence length) with the global condition \(c_g\) as follows:
\begin{equation}
    \text{Gating}_g(y, c_g) = y \cdot \mathbf{1}\mathbf{W}_g(c_g).
\label{eq:gate_global}
\end{equation}
Here, \(\mathbf{W}_g\in\mathbb{R}^{h\times h}\) is an MLP, and $\mathbf{1}$ is a $L\times1$ all-ones matrix, which ``broadcasts'' the scales to match the sequence length $L$ -- such that the same condition is applied to all tokens in the sequence (hence global).

Similarly, for per-token gating layers, the frame-aligned conditioning \(c_f\) is injected into the audio stream for precise feature modulation via
\begin{equation}
    \text{Gating}_f(y, c_f) = y \cdot \mathbf{W}_f(c_f), 
\label{eq:gate_token}
\end{equation}
where \(\mathbf{W}_f\in\mathbb{R}^{h\times h}\) is an MLP.
Different from \Cref{eq:gate_global}, the scales are applied per token without broadcasting, 

\subsection{Details on Synchronization Features}
\begin{figure}
    \centering
    \includegraphics[width=0.4\linewidth]{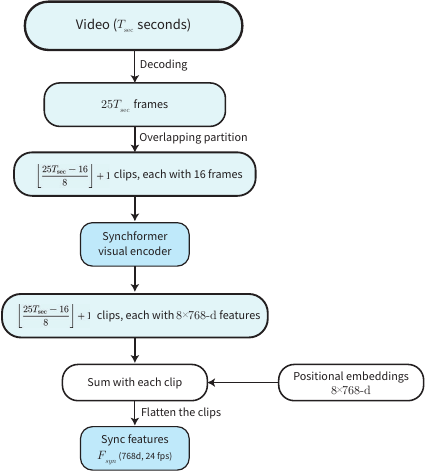}
    \vspace{1ex}
    \caption{Synchformer feature extraction.}
    \label{app:fig:synchformer-details}
\end{figure}

We use the visual encoder of Synchformer~\cite{iashin2024synchformer} to extract synchronization features. 
We use the pretrained audio-visual synchronization model trained on AudioSet, provided by~\citet{iashin2024synchformer}.
As input, we obtain frames at 25 fps.  
Synchformer partitions these frames into overlapping clips of 16 frames with stride 8 and produces features of length 8 for each clip.
Thus, for a video of length \(T_\text{sec}\) seconds, the sequence length of the synchronization features is
\begin{equation}
    L_\text{sync} = 8\left(\left\lfloor{
    \frac{25T_\text{sec} - 16}{8}
    }\right\rfloor + 1\right).
\end{equation}
The corresponding feature fps is
\begin{equation}
    \text{FPS}_\text{sync} = \frac{L_\text{text}}{T_\text{sec}}.
\end{equation}
In this paper, we experimented with \(T_\text{sec}=8\)  and \(T_\text{sec}=10\).
In both cases, \(\text{FPS}_\text{sync}\) is exactly 24. 
Additionally, we introduce a learnable positional embedding of length 8 (matching the number of features in each clip processed by Synchformer) that is added to the Synchformer features, as illustrated in~\Cref{app:fig:synchformer-details}.

\subsection{Illustration of the ``sum sync with visual'' Ablation}\label{app:sec:sum-sync}
\begin{figure}
    \centering
    \includegraphics[width=0.7\linewidth]{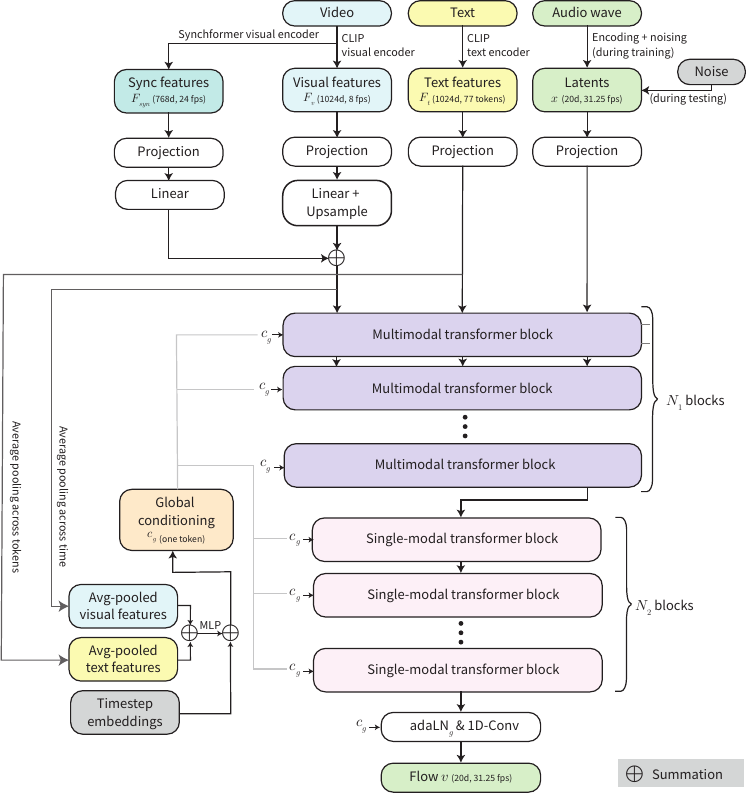}
    \vspace{1ex}
    \caption{Illustration of the ``sum sync with visual'' ablation.}
    \label{fig:sum_sync_with_visual}
\end{figure}

\Cref{fig:sum_sync_with_visual} illustrates the network architecture for  the ``sum sync with visual'' ablation in the ``conditional synchronization module'' paragraph.
The visual features are upsampled using the nearest neighbor to match the frame rate of the synchronization features. 
This architecture has a worse \fdpasst{}, IB-score, synchronization (DeSync) but a better inception score (IS), which we hypothesize is due to the increased number of visual tokens in the upsampling step, leading to finer-grained computations.

\subsection{Visualization of Aligned RoPE}\label{app:sec:aligned-rope}
\begin{figure}[ht]
    \centering
    \includegraphics[width=0.49\linewidth]{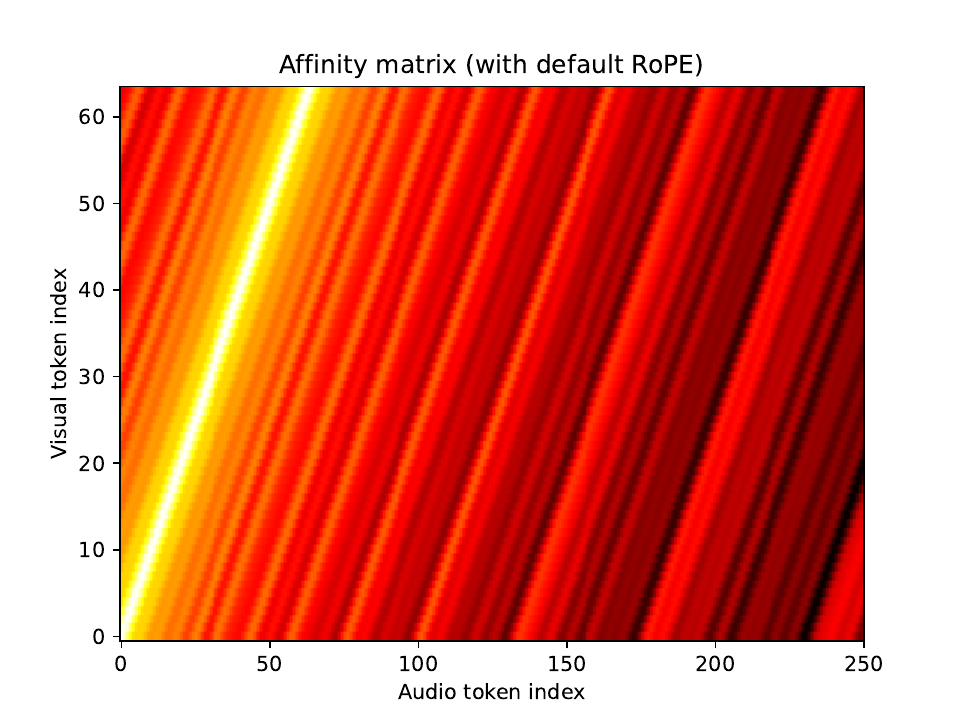}
    \includegraphics[width=0.49\linewidth]{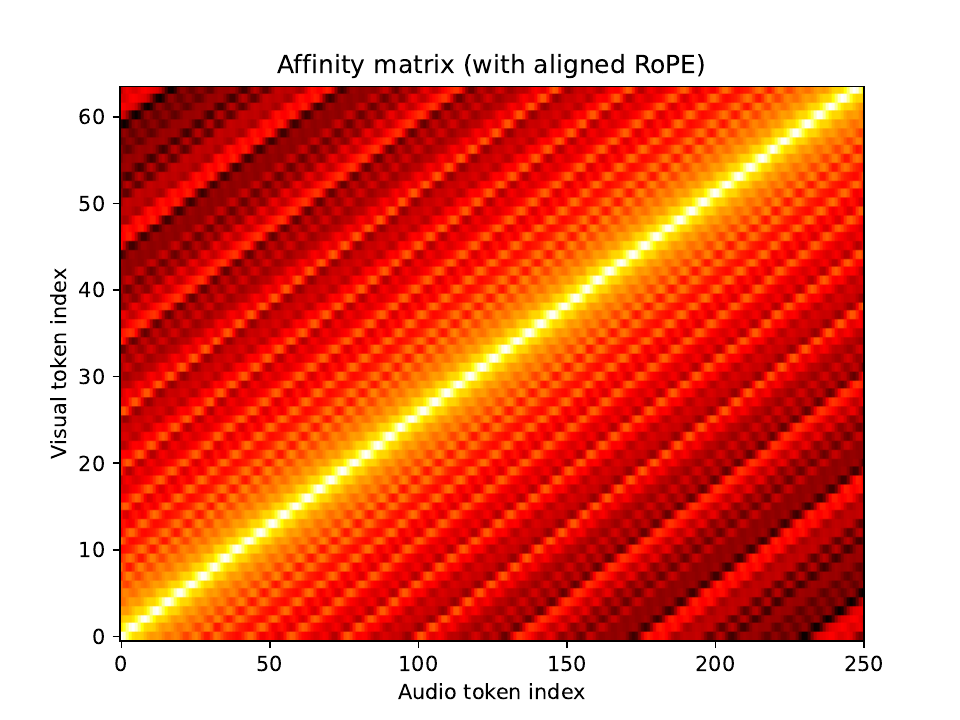}
    \caption{Affinity visualizations between two sequences with different frame rates when default/aligned RoPE embeddings are used. 
    \textbf{Left}: with default RoPE, the sequences are not aligned.
    \textbf{Right}: with our proposed aligned RoPE, we attain temporal alignment.
    }
    \label{app:fig:aligned_rope}
\end{figure}

To visualize the effects of using aligned RoPE~\cite{su2024roformer}, we compare the dot-product affinity of two sequences \(\mathbf{1}^{250\times C}\) and \(\mathbf{1}^{64\times C}\) when RoPE is applied.
Here, \(250\) represents the audio sequence length (31.25 fps for 8 seconds),  \(64\) represents the visual sequence length (8 fps for 8 seconds), and \(C=64\) is the channel size.
Concretely, we visualize
\begin{equation}
    \text{RoPE}_\text{default}(\mathbf{1}^{250\times C}) \left(\text{RoPE}_\text{default}(\mathbf{1}^{64\times C})\right)^T, 
\end{equation}
and, 
\begin{equation}
    \text{RoPE}_\text{aligned}(\mathbf{1}^{250\times C}) \left(\text{RoPE}_\text{aligned}(\mathbf{1}^{64\times C})\right)^T, 
\end{equation}
in~\Cref{app:fig:aligned_rope}.
Temporal alignment is attained when we use aligned RoPE.

\section{Training Details}\label{app:sec:training}

\paragraph{Training setup.}
Unless otherwise specified, we used the same set of hyperparameters for all model sizes.
To train the models, we use the base learning rate of \(1\mathrm{e}\)-\(4\), with a linear warm-up schedule of 1K steps,  for 300K iterations, and with a batch size of 512.
We use the AdamW optimizer~\cite{kingma2014adam,loshchilov2017decoupled} with \(\beta_1=0.9\), \(\beta_2=0.95\), and a weight decay of \(1\mathrm{e}\)-\(6\).
If the default \(\beta_2=0.999\) was used instead, we notice occasional training collapse (to NaN).
For learning rate scheduling, we reduce the learning rate to \(1\mathrm{e}\)-\(5\) after 80\% of the training steps, and once again to \(1\mathrm{e}\)-\(6\) after 90\% of the training steps.
For model exponential moving average (EMA), we use the post-hoc EMA~\cite{karras2024analyzing} formulation with a relative width \(\sigma_{\text{rel}}=0.05\) for all models.
For training efficiency, we use {\tt bf16} mixed precision training, and all the audio latents and visual embeddings are precomputed offline and loaded during training.
\Cref{tab:training_res} summarizes the training resources we used for each model size.

\begin{table}[ht]
\vspace{1ex}
    \centering
    \begin{tabular}{lccc}
    \toprule
    Model & Number of GPUs used & Number of hours to train & Total GPU-hours \\
    \midrule
    MMAudio-S-16kHz & 2 & 22 & 44 \\
    MMAudio-S-44.1kHz & 2 & 26 & 52 \\
    MMAudio-M-44.1kHz & 8 & 21 & 168 \\
    MMAudio-L-44.1kHz & 8 & 38 & 304 \\
    \midrule
    \bottomrule
    \end{tabular}
    \vspace{1ex}
    \caption{
    The amount of training resources used for each model size.
    H100 GPUs are used in all settings.
    }
    \label{tab:training_res}
\vspace{1ex}
\end{table}

\paragraph{Balancing multimodal training data.}
Since we have significantly more audio-text training data (951K) than audio-text-visual data (180K), we balance the dataset by duplicating the audio-text-visual samples before random shuffling in each epoch. 
By default, we apply a 5X duplication for a rough 1:1 data sampling ratio. 
For the ``medium'' and ``large'' models, we reduce the duplication ratio to 3X to mitigate overfitting.

\paragraph{Duplicated videos.}
We observe VGGSound dataset~\cite{chen2020vggsound} contains duplicated videos, likely due to multiple uploads of the same video to YouTube under different video IDs.
For instance, videos {\tt 4PjEi5fFD6A} (in training set) and {\tt FhaYvI1yrUM} (in test set) are the same video.\footnote{Other uploads of this video that are \emph{not} part of the VGGSound dataset include {\tt 1MQkMdlBezY} and {\tt vHmRikW9axQ}.}
In \Cref{app:sec:overlap}, we remove train-test sets overlaps by comparing the video IDs, though this method does not eliminate repeated uploads.
Since prior works have been trained on the same dataset, our training scheme remains a fair comparison.

\section{Additional Visualizations}

\begin{figure}[ht]
    \centering
    \includegraphics[width=\linewidth]{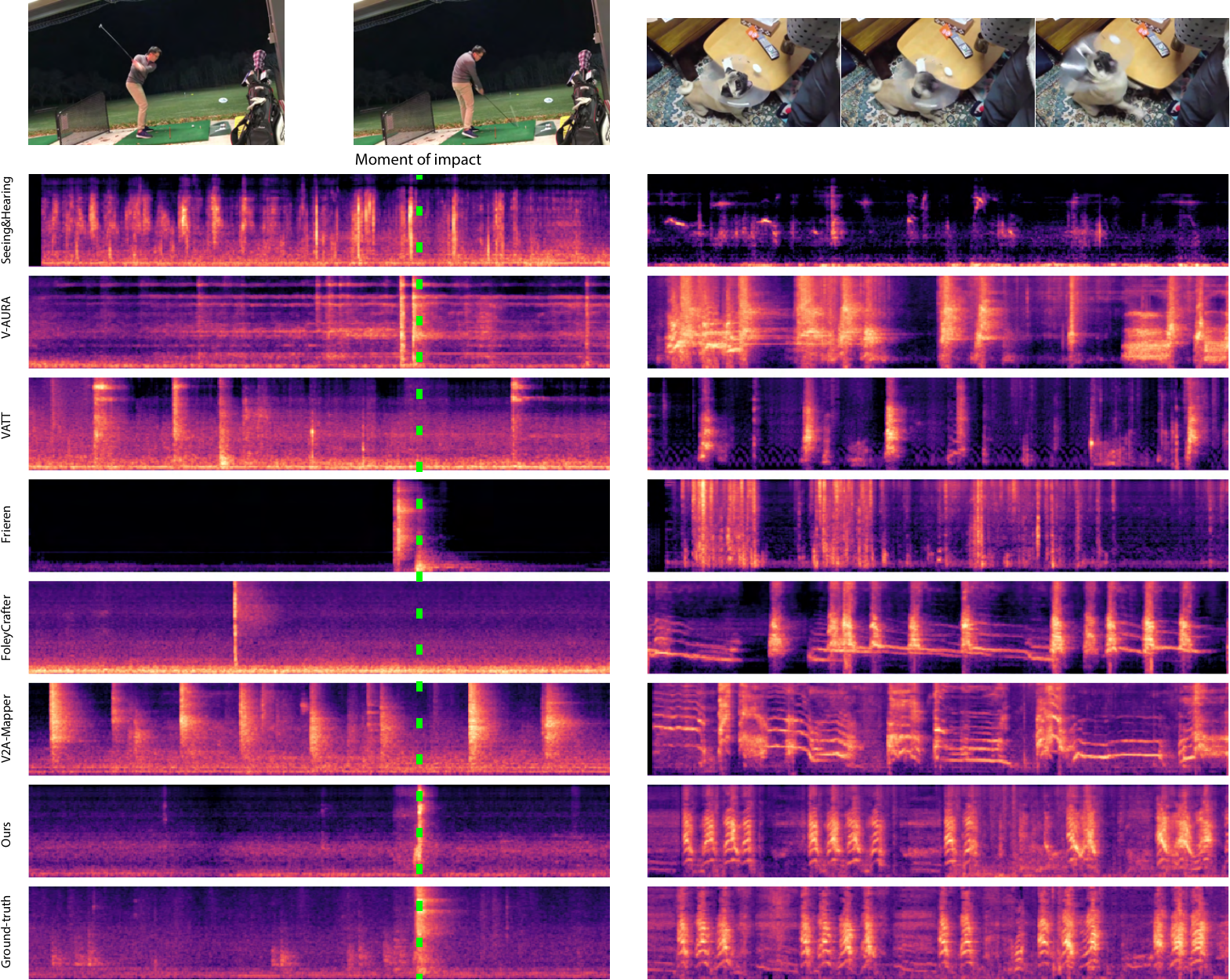}
    \caption{
    Left: our method can precisely capture the distinct audio event of striking a golf ball.
    Right: a dog barks in successive bursts. 
    Our generation does not line up with the ground-truth as precisely due to the ambiguous nature of video-to-audio generation, but does capture the rapid bursts.
    }
    \label{fig:spectrogram_vis_2}
\end{figure}

\begin{figure}[ht]
    \centering
    \includegraphics[width=\linewidth]{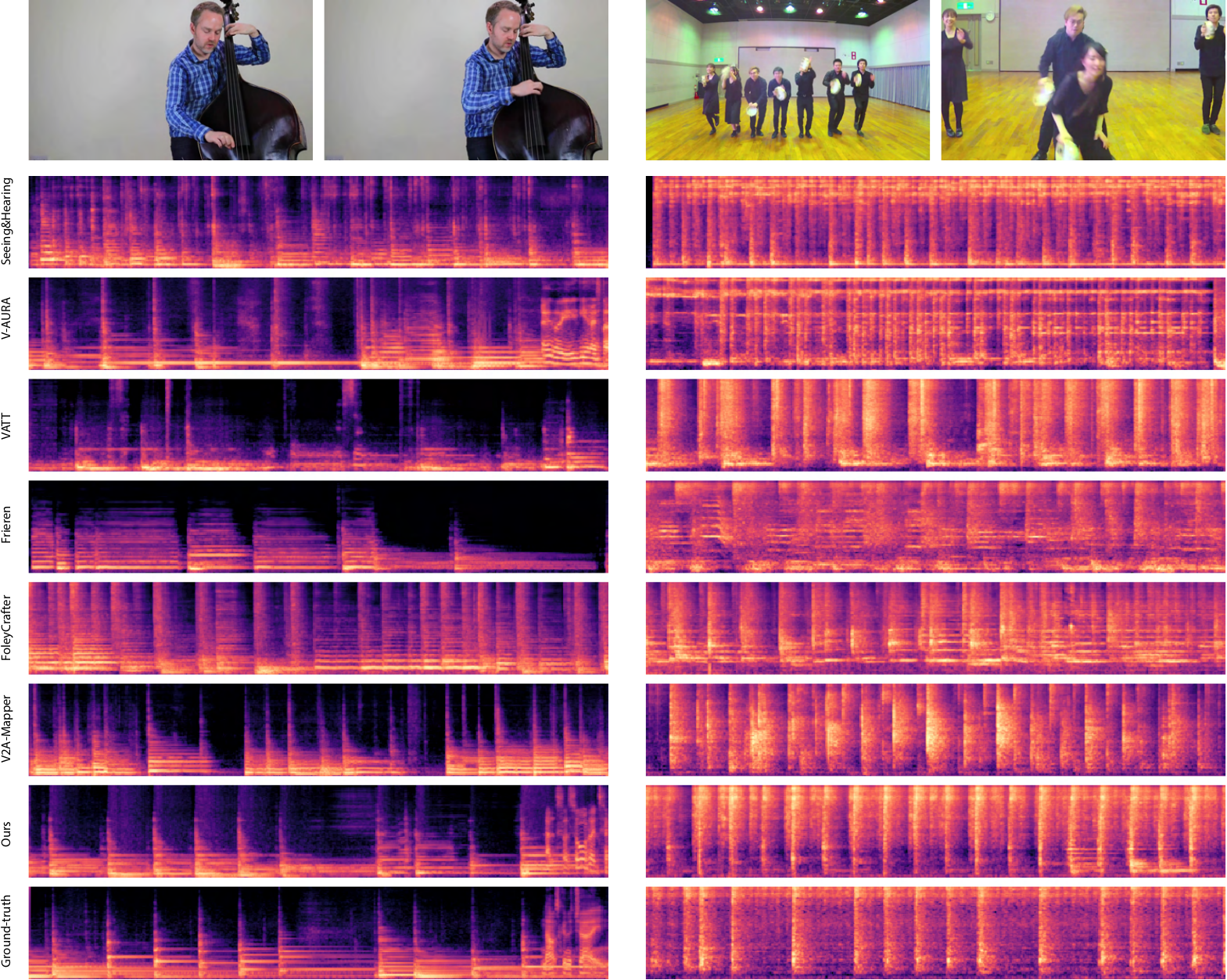}
    \caption{
    Left: when visible audio events (\eg, when a string is played) can be clearly seen, MMAudio captures them much more precisely than existing methods.
    Right: in a complex scenario, MMAudio does not always generate audio aligned to the ground-truth (as common in the generative setting) but the generation is often still plausible.
    }
    \label{fig:spectrogram_vis_3}
\end{figure}

\begin{figure}[ht]
    \centering
    \includegraphics[width=\linewidth]{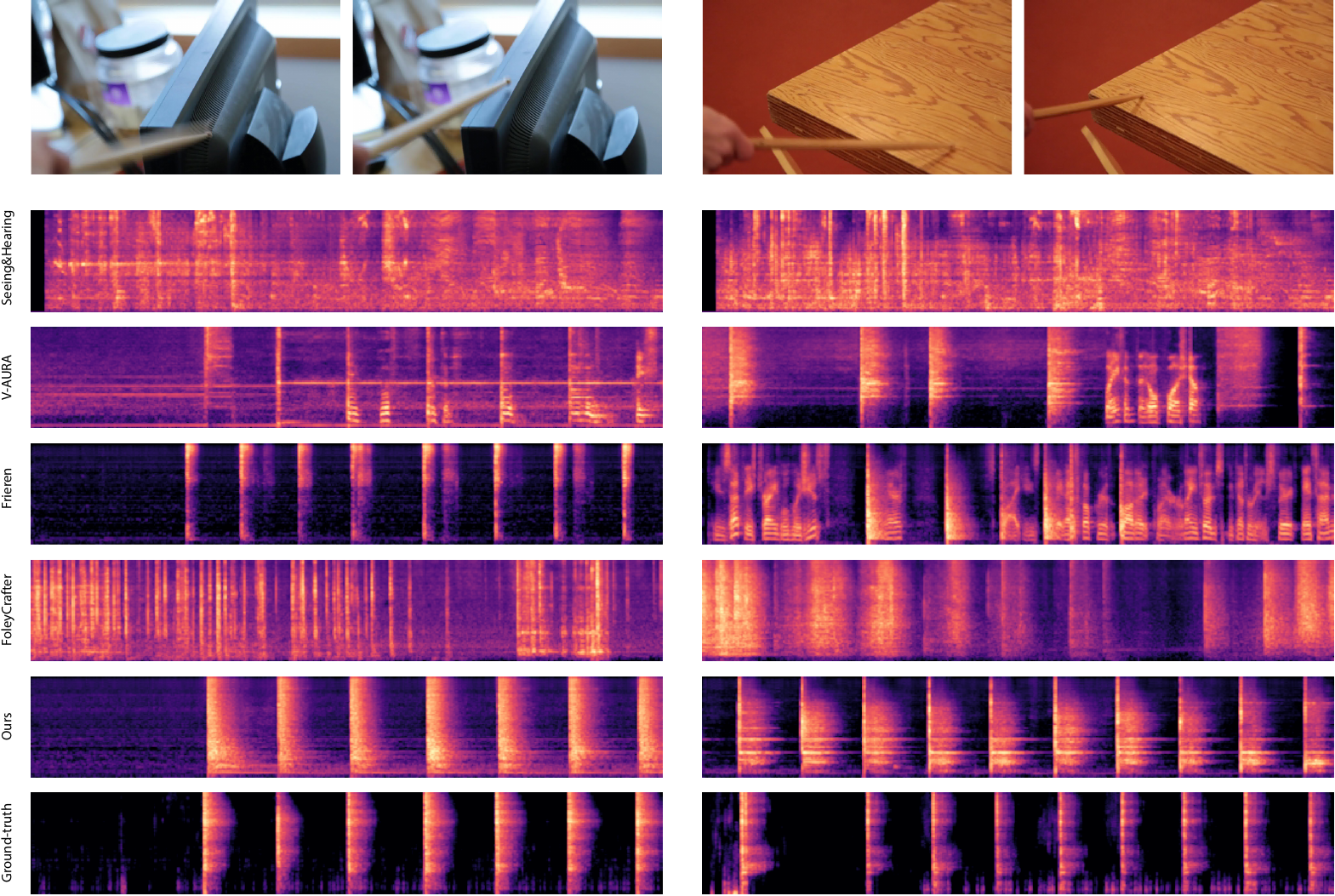}
    \caption{
    Comparisons of prior works with MMAudio on the Greatest Hits~\cite{owens2016visually} dataset.
    }
    \label{fig:spectrogram_vis_greatesthits}
\end{figure}

We provide generated samples and comparisons with state-of-the-art methods on our project page \url{https://hkchengrex.com/MMAudio/video_main.html}.
Below, we provide additional spectrogram visualizations comparing our method with prior works in \Cref{fig:spectrogram_vis_2,fig:spectrogram_vis_3,fig:spectrogram_vis_greatesthits}.

%% file: tab/tab-user-study.tex
\begin{table}[h]
\vspace{1ex}
    \centering
    \begin{NiceTabular}{lccc}
    \toprule
    Method & Audio quality\(\uparrow\) & Semantic alignment\(\uparrow\) & Temporal alignment\(\uparrow\) \\
    \midrule
    Seeing\&Hearing~\cite{xing2024seeing} & 2.65\pmnum{1.05} & 3.10\pmnum{1.24} &	1.85\pmnum{0.99} \\
    V-AURA~\cite{viertola2024temporally} & 3.59\pmnum{1.02} & 3.70\pmnum{1.17} & 3.65\pmnum{1.16} \\
    VATT~\cite{liu2024tellhearvideo} & 2.66\pmnum{0.99} & 3.32\pmnum{1.17} & 2.04\pmnum{1.07} \\
    V2A-Mapper~\cite{wang2024v2a} & 3.00\pmnum{0.95} & 3.28\pmnum{1.27} & 2.03\pmnum{1.11} \\
    \rowcolor{defaultColor}
    MMAudio-L-44.1kHz & \textbf{4.14\pmnum{0.77}} & \textbf{4.52\pmnum{0.74}} & \textbf{4.46\pmnum{0.80}} \\
    \midrule
    \bottomrule
    \end{NiceTabular}
    \vspace{1ex}
    \caption{
    Average ratings for each method from the user study.
    We show mean\(\pm\)std in each aspect.
    }
    \label{tab:user_study}
    \vspace{1ex}
\end{table}

%% file: tab/tab-moviegen.tex
\begin{table}[h]
\vspace{1ex}
\small
    \centering
    \begin{NiceTabular}{l@{\hspace{6pt}}c@{\hspace{6pt}}c@{\hspace{6pt}}c@{\hspace{6pt}}c@{\hspace{6pt}}c@{\hspace{6pt}}c@{\hspace{6pt}}c@{\hspace{6pt}}c@{\hspace{6pt}}c}
    \toprule
    & & & \multicolumn{3}{c}{Subjective metrics} & \multicolumn{4}{c}{Objective metrics} \\
    \cmidrule(lr{\dimexpr 4\tabcolsep-16pt}){4-6}
    \cmidrule(lr{\dimexpr 4\tabcolsep-16pt}){7-10}
    \footnotesize Method & 
    \footnotesize Param & 
    \footnotesize Training data & 
    \footnotesize Audio qual.\(\uparrow\) & 
    \footnotesize Semantic align.\(\uparrow\) & 
    \footnotesize Temporal align.\(\uparrow\) & 
    \footnotesize IS\(\uparrow\) & 
    \footnotesize IB-score\(\uparrow\) & 
    \footnotesize CLAP\(\uparrow\) & 
    \footnotesize DeSync\(\downarrow\) \\
    \midrule
    Movie Gen Audio~\cite{polyak2024movie} & 13B & \(\mathcal{O}(1,000,000)\)h &
    \textbf{3.93\pmnum{0.92}} & \textbf{4.36\pmnum{0.74}} & 3.52\pmnum{1.21} & \textbf{8.40} & \textbf{36.26} & \textbf{0.4409} & 1.006 \\
    \rowcolor{defaultColor}
    MMAudio-L-44.1kHz & 1.03B & \(\sim8,200\)h &
    \textbf{3.93\pmnum{0.89}} & 4.26\pmnum{0.71} & \textbf{3.62\pmnum{1.03}} & \textbf{8.40} & 27.01 & 0.4324 & \textbf{0.771} \\
    \midrule
    \bottomrule
    \end{NiceTabular}
    \vspace{1ex}
    \caption{
    Comparisons between Movie Gen Audio and MMAudio in both subjective metrics (from user study) and objective metrics.
    For the subjective metrics, we show mean\(\pm\)std.
    }
    \label{tab:moviegen}
\vspace{1ex}
\end{table}

%% file: tab/tab-overlap.tex
\begin{table}[ht]
\vspace{1ex}
\small
    \centering
    \begin{NiceTabular}{lcccc}
    \toprule
    Test sets (number of samples) & \multicolumn{4}{c}{{Training sets}} \\
    \cmidrule(lr{\dimexpr 4\tabcolsep-16pt}){2-5}
    & AudioSet & AudioCaps & VGGSound & AVSync15 \\
    \midrule
    AudioCaps (975) & 580 (59.5\%) & - & 147 (15.1\%) & - \\
    VGGSound (15,496) & 132 (0.9\%) & 13 (0.1\%) & - & 59 (0.4\%) \\
    AVSync-15 (150) & - & - & 144 (96.0\%) & - \\
    \midrule
    \bottomrule
    \end{NiceTabular}
    \vspace{1ex}
    \caption{
    Overlaps between training and test sets of different datasets.
    The percentage denotes the proportion of overlapping data in the entire test set.
    ``-'' means that we did not compute this data (we do not train or test on AVSync15).
    }
    \label{tab:data_overlap}
\vspace{1ex}
\end{table}